%% file: main.tex
\documentclass[acmtog,authorversion,nonacm]{acmart}
\usepackage{amsmath}
\usepackage[ruled]{algorithm2e} 

\usepackage{caption}
\usepackage{tabularx}
\usepackage{multicol,lipsum}
\acmSubmissionID{450}

\AtBeginDocument{%
  \providecommand\BibTeX{{%
    \normalfont B\kern-0.5em{\scshape i\kern-0.25em b}\kern-0.8em\TeX}}}

\acmConference[Conference acronym 'XX]{Make sure to enter the correct
  conference title from your rights confirmation emai}{June 03--05,
  2018}{Woodstock, NY}

\acmPrice{15.00}
\acmISBN{978-1-4503-XXXX-X/18/06}

\input{src/macros}

\begin{document}
\title{DeepTree: Modeling Trees with Situated Latents}

\author{Xiaochen Zhou}
\affiliation{%
  \institution{Purdue University}
  \country{USA}
  }
\email{zhou1178@purdue.edu}

\author{Bosheng Li}
\email{li2343@purdue.edu}
\affiliation{%
  \institution{Purdue University}
  \country{USA}
}

\author{Bedrich Benes}
\affiliation{%
  \institution{Purdue University}
  \country{USA}
  }
\email{bbenes@purdue.edu}

\author{Songlin Fei}
\affiliation{%
  \institution{Purdue University}
  \country{USA}
  }

\author{S\"oren Pirk}
\affiliation{%
  \institution{Adobe Research}
  \country{USA}
  }

\renewcommand{\shortauthors}{Zhou, et al.}

\input{src/00-abstract}

\begin{teaserfigure}
  \vspace{-2mm}
  \includegraphics[width=\textwidth]{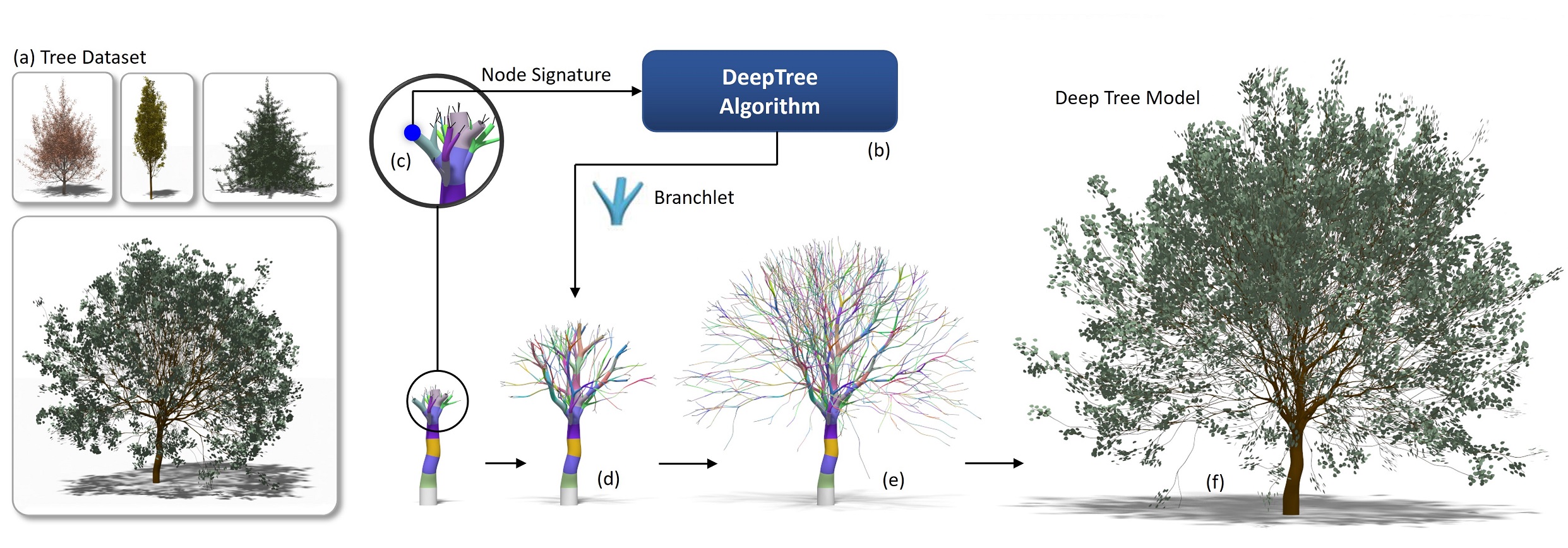}
  \vspace{-4mm}
  \caption{Illustration of our DeepTree (\name) generation approach: we use a common procedural model to generate a dataset of different species (a). As part of our algorithm, we train a neural network pipeline (b) to predict the topological and geometric properties of branching patterns based on a situated latent space. This means that branch features are predicted locally and by only considering a single node (c). Given a node, our network pipeline is able to predict branchlets, the branching structures of the immediate successors of a node -- thereby mimicking the developmental process of a tree model (d, e). Eventually, our algorithm will automatically terminate when a complex tree model has been generated (f). Branchlets are shown with colors in (c), (d), and (e).  }
  \vspace{4mm}
\label{fig:teaser}
\end{teaserfigure}

\maketitle

\input{src/01-introduction}
\input{src/02-related_work}
\input{src/03-overview}
\input{src/04-method}

\input{src/05-results}
\input{src/06-conclusion}

\begin{acks}
We want to thank Liangliang Nan for providing a point cloud for the tree model from Fig.~\ref{fig:reconstruction_real}.
\end{acks}

\bibliographystyle{ACM-Reference-Format}
\bibliography{main}

\input{src/99-appendix}

\end{document}

%% file: src/macros.tex
\usepackage{color}
\usepackage{soul}
\usepackage{multirow}
\usepackage{rotating}
\usepackage{amsmath}
\usepackage{algorithmicx}
\usepackage{wrapfig}
\usepackage[normalem]{ulem}

\definecolor{turquoise}{cmyk}{0.65,0,0.1,0.1}
\definecolor{purple}{rgb}{0.65,0,0.65}
\definecolor{darkgreen}{rgb}{0.0, 0.5, 0.0}
\definecolor{darkred}{rgb}{0.5, 0.0, 0.0}
\definecolor{darkblue}{rgb}{0.0, 0.0, 0.5}
\definecolor{blue}{rgb}{0.0, 0.0, 1.0}
\definecolor{orange}{rgb}{1.0, 0.5, 0.0}

\newcommand{\name}{\textit{DeepTree}}

\newcommand{\erase}[1]{}

\newcommand{\hide}[1]{{}}
\newcommand{\ch}[1]{{\color{black}#1}}
\newcommand{\chh}[1]{{\color{black}#1}}

\definecolor{darkblue}{rgb}{0.0, 0.2, 0.5}

\usepackage{marginnote}

%% file: src/00-abstract.tex
\begin{abstract}
In this paper, we propose \textit{DeepTree}, a novel method for modeling trees based on learning developmental rules for branching structures instead of manually defining them. We call our deep neural model ``situated latent'' because its behavior is determined by the intrinsic state -encoded as a latent space of a deep neural model- and by the extrinsic (environmental) data that is ``situated'' as the location in the 3D space and on the tree structure. We use a neural network pipeline to train a situated latent space that allows us to locally predict branch growth only based on a single node in the branch graph of a tree model. We use this representation to progressively develop new branch nodes, thereby mimicking the growth process of trees. Starting from a root node, a tree is generated by iteratively querying the neural network on the newly added nodes resulting in the branching structure of the whole tree. Our method enables generating a wide variety of tree shapes without the need to define intricate parameters that control their growth and behavior. Furthermore, we show that the situated latents can also be used to encode the environmental response of tree models, e.g., when trees grow next to obstacles. We validate the effectiveness of our method by measuring the similarity of our tree models and by procedurally generated ones based on a number of established metrics for tree form. 
\end{abstract}

\ccsdesc[500]{Computing methodologies~Generative and developmental approaches}
\ccsdesc[500]{Computing methodologies~Shape analysis}
\ccsdesc[300]{Computing methodologies~Computer vision problems}

\keywords{Botanical Tree Models, Deep Learning, Shape Modeling, Generative Methods}

%% file: src/01-introduction.tex
\section{Introduction}\label{sec:intro}
Vegetation is ubiquitous in almost all environments, ranging from vast outdoor landscapes to indoor spaces. Consequently, models of trees and plants serve as essential assets in applications such as games and movies, mixed reality, architecture and urban planning, agriculture, and forestry, or even the training of autonomous agents. For many applications, it is of utmost importance to define plant models with a high degree of geometric detail to produce high-quality renderings or model interactions with plant models, e.g., to plan a path around them. Procedural models have proven to effectively capture the wide variety of plant forms found in Nature and efficiently generate large sets of models.

Procedural and developmental modeling of vegetation in CG has been used in diverse ways, often combined with other approaches, such as modeling the environmental response of trees~\cite{Palubicki:2009:STM}, the reconstruction of tree models from images or point sets~\cite{Li2021ToG,Livny:2011:Sigg,du2019adtree}, inverse procedural modeling~\cite{Stava2014}, by focusing on interactive modeling with user-defined sketches~\cite{Longay:SBIM:12}, or by leveraging neural networks that compose trees of procedurally generated branching patterns~\cite{Liu2021ToG}. The breadth of these approaches is a testament to the complexity of realistically modeling plants. However, despite the progress, generating trees with procedural models remains an open and challenging research problem.

Our key inspiration comes from Nature's ability to compress and encode structural and behavioral properties into a highly compact form, such as DNA. Bud development is intrinsically determined by the DNA, and extrinsically by the environment (light, gravity, nutrients), and the position in the tree hierarchy~\cite{acquaah2009principles,darcy}. The bud development is a response to these factors; it can grow to a new location, create lateral buds, become dormant etc. 
The ability of the DNA to encode complex interactions that lead to shape as an emergent 3D structure (gene expression) is one of the key open problems in science~\cite{khatri2005ontological}. Inspired by this problem, we seek to develop a generative model that would capture the resulting shape of the behavior. While engineering such behavior is challenging because each tree node experiences varying conditions, deep learning is suitable for learning and encoding the bud's response to such conditions. 

We propose a novel deep neural model that we call ``situated latent'' because its behavior is determined by the intrinsic state -encoded as a latent space of a deep neural model- and by the extrinsic (environmental) data that is ``situated'' as the location in the 3D space and on the tree structure. We use a large dataset of 3D tree models to train a pipeline of neural networks on \textit{branchlets} -- atomic branching structures consisting of a node and its immediate children. We learn a situated latent space, a representation that makes predictions for branch growth based on where a branch node is located in the growth space of a tree model. Once trained, the neural networks can predict the topology (number of children) and geometry (position and the branch width of children) for a single input node. The deep neural model is a  compact representation that encodes the tree response based on the environment (extrinsic) and the positional information within the tree structure (intrinsic). Similar to the DNA, we encode the entire tree and its environmental response as a single compact deep neural representation.

A new tree is generated iteratively by executing the network on individual nodes of the branching structure. Starting from the root node, the single network is executed for each node, and it predicts the number and positions of its children. The predicted child nodes are added to the branch graph, and in the next iteration, the network predicts their immediate children again. Similar to procedural modeling algorithms that are also implemented recursively or iteratively, this allows producing complex branching structures only with a few iterations. While \name\ learns the response of branchlets, it is \textit{not} a reconstruction method as the generated trees are non-deterministic and only share low-level common properties of the entire training set.

Our results show that the situated latent space can encode and generate diverse branching patterns of multiple tree species by successfully predicting branch growth's primary and secondary attributes (e.g., branching angles, internode length, branch width, etc.). The network also predicts the termination of branches (no immediate children) for twigs toward the outer regions of the tree crown, which then terminates the development of a model. Moreover, we show that the situated latent space can also be used to encode the environmental response of trees, i.e., the adaptation of branching structures according to tropisms and obstacles in the model's environment.  

We are not aware of any deep neural models capable of generating 3D biological tree shape. Therefore, we validate the effectiveness of our method in encoding tree form by comparing tree models generated with a state-of-the-art procedural model, and our qualitative results show that our method generates branching structures with almost identical geometric properties. We measure the similarity of branching structures based on histograms over the geometric properties at different levels of fully developed tree models. We also use a recently proposed perceptual metric to assess the quality of generated tree models further. 

Examples of a procedurally generated tree and a \name\ model are shown in Fig.~\ref{fig:teaser}. The \name\ model is generated by iteratively querying a neural network on the current set of terminal nodes. The network predicts new branch nodes -- thereby mimicking the growth process -- that are then added to the branch graph. A complex tree model can be generated with a few iterations. 

In summary, our contributions are: (1)~we propose a novel method for learning to predict the topological and geometric properties of branchlets based on a novel neural network pipeline; (2)~we show that the learned situated latent space can be used to iteratively generate tree models, which provides a novel way to encode and generate tree form; (3)~we validate our method through state-of-the-art qualitative and quantitative metrics.  

%% file: src/02-related_work.tex
\vspace{-2mm}
\section{Related Work}

\begin{figure*}[t]
\centering
\includegraphics[width=1\textwidth]{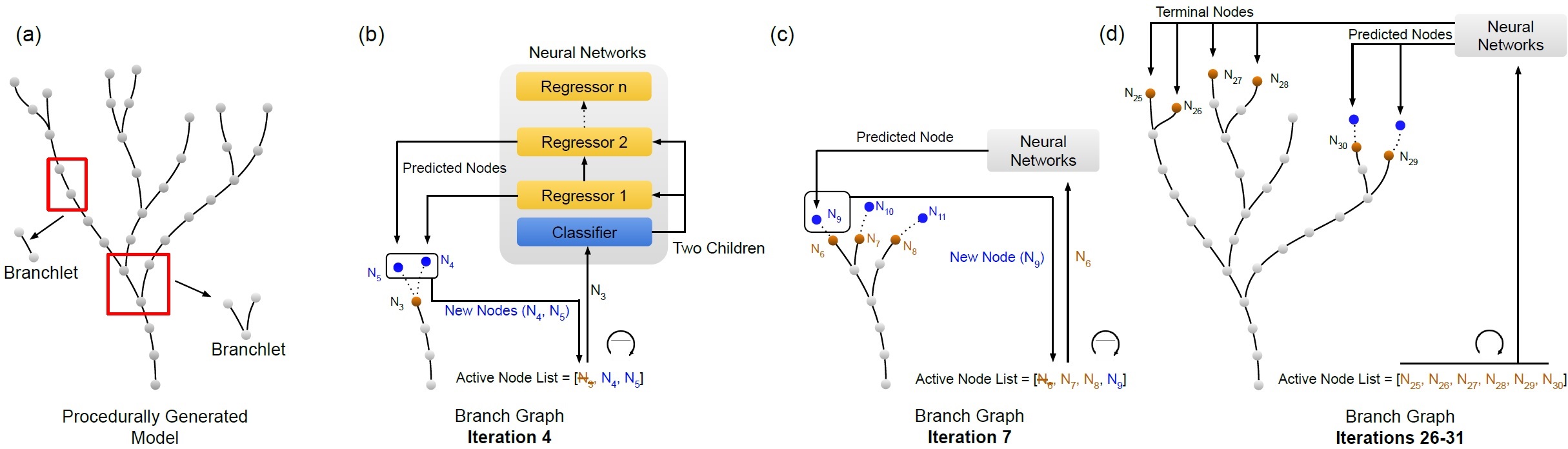}
\vspace{-7mm}
\caption{Overview: We use branchlets of tree models generated with a common procedural model to train our neural networks (a). To generate a tree, we iteratively use our neural network pipeline to predict the topology and geometry of new nodes. A classification network predicts the number of new children for a node that triggers a cascade of regression networks predicting the signatures of each child by using the output of the previous regression network (e.g., Regressor 2 uses the parent node signature as well as the output signature of Regressor 1). Starting with the root node, we maintain a list of active nodes. We pop an active node from the list in each iteration and trigger the neural networks to predict its children. The children are then added to the branch graph and the list of active nodes (b, c). The generation of a tree model stops when the classification network predicts zero children for an active node, which happens individually for different branches (d).}
\vspace{-3mm}
\label{fig:overview}
\end{figure*}

Due to the importance of vegetation in several application domains, modeling trees and plants has received significant research attention over the past decades (see review~\cite{Pirk:2016:MPL:2897826.2927332}). The goal of the early approaches was to focus on faithfully modeling the morphology of branching structures. Approaches include fractals~\cite{Aono:CGaA:84}, repetitive patterns~\cite{Oppenheimer:86}, environmentally sensitive automata~\cite{Greene:89:SIGG}, and particle-systems~\cite{10.1145/325165.325250}. L-systems~\cite{Pwp90BOOK} -- as one of the most fundamental procedural approaches -- enable to define the tree growth as a set of parallel production rules and has been shown to allow for capturing a wide variety of plant forms, even when considering the environmental response of plants~\cite{Mech:96:SIGG}. However, while L-systems are a powerful mechanism for generating plant form, defining production rules to generate complex trees is an elaborate task that requires a substantial amount of knowledge, which often renders their use for applications impractical. Early procedural models~\cite{weber-1995} aim to address these shortcomings by combining rule-based approaches with geometric modeling and user interaction~\cite{Lintermann:99:CGaA}. However, they are more specialized and do not support modeling the same range of branching patterns. Finally, the goal of inverse procedural modeling~\cite{Stava2014,Talton:11:ToG} is to automatically find the parameter values of procedural models of plants through optimization schemes or to find the production rules~\cite{Guo2020InverseProceduralModeling}. 

Recent procedural methods for plants emphasize more principled representations~\cite{https://doi.org/10.1111/cgf.12744,10.1007/s00607-014-0424-7} or the underlying biological processes of plant development. \citeauthor{Stava2014}~\shortcite{Stava2014}, for example, add biological priors to their procedural model to provide more nuanced control for shape-defining parameters. Other methods focus on the self-organization of plants~\cite{Palubicki:2009:STM}, their environmental response~\cite{Mech:96:SIGG,Pirk:2012:PTI:2185520.2185546}, the competition for resources~\cite{Greene:89:SIGG,10.5555/791218.791582}, or modeling features of specific types of plants~\cite{CGF:CGF12736}. The availability of more powerful graphics hardware-enabled approaches that use procedural models of plants to simulate the animation of growth~\cite{Pirk:2012:CAM:2366145.2366188,Haedrich:2017}, their physical response to wind~\cite{Pirk:2014:WTC:2661229.2661252,Habel_09_PGT} and fire~\cite{Pirk:2017:IWC:3130800.3130814}, advanced material properties of wood~\cite{Wang:2017:BMB,Zhao:2013:IAS:2461912.2461961}, or to find efficient representations of tree models~\cite{Neubert.etal:2011:IMVDPLBS}. \chh{More recently, module-based tree representations -similar to our branchlets- have gained popularity to enable simulating complex vegetation-related phenomena, such as wildfires~\cite{10.1145/3450626.3459954} or climatic gradients~\cite{10.1145/3528223.3530146}.}

Sketch-based and data-driven approaches often use procedural models by guiding the generation according to the defined input data to generate convincing branching structures. User-defined sketches provide the intriguing advantage that plants can be modeled with direct control to meet artistic requirements~\cite{Ijiri:EG:06,OkabeSketchTree07,wither:hal-00366289,10.1145/1409060.1409062}. \citeauthor{Longay:SBIM:12}~\shortcite{Longay:SBIM:12} propose an advanced framework to efficiently produce complex tree models that can even run on mobile hardware. Data-driven approaches, on the other hand, focus on solving a reconstruction task to generate plant models to faithfully match a captured tree that can either be represented as one or several images~\cite{Bradley:2013:IRS:2461912.2461952,Tan:2007:ITM:1276377.1276486,Tan:2008:SIT:1409060.1409061,NeubertTreeParticle07}, videos~\cite{Li:2011:MGM:2070781.2024161}, or point clouds~\cite{Livny:2011:Sigg,Xu07}. More recently, it has also been recognized that neural networks are a powerful means to guide procedural modeling by learning bounding volumes to aid the reconstruction from single images~\cite{Li2021ToG} by decomposing point clouds~\cite{Liu2021ToG} into semantically meaningful patterns, or by learning parameters for the placement of plants~\cite{10.1145/3502220}. Moreover, synthesizing large data collections from a few examples appears to be an increasingly important pursuit~\cite{https://doi.org/10.1111/cgf.13752}. Finally, some approaches also focus on defining trees based on partially defined branching structures -- procedurally generated -- to enable a level of detail schemes~\cite{https://doi.org/10.1111/cgf.13088} or the processing of large collections of plants~\cite{10.1145/3306346.3323039}. 

\chh{The work of \citeauthor{Estrada2015-mh}~\shortcite{Estrada2015-mh} follows a similar objective to ours. Their approach also uses a  parametric tree-growth model to regularize topology estimation for tree-like structures. However, despite these advances, none of the existing approaches for trees and plants uses neural networks to encode the rules and parameters for locally encoding branching patterns for botanical tree models.}

%% file: src/03-overview.tex
\begin{figure*}[t]
\centering
\includegraphics[width=1\textwidth]{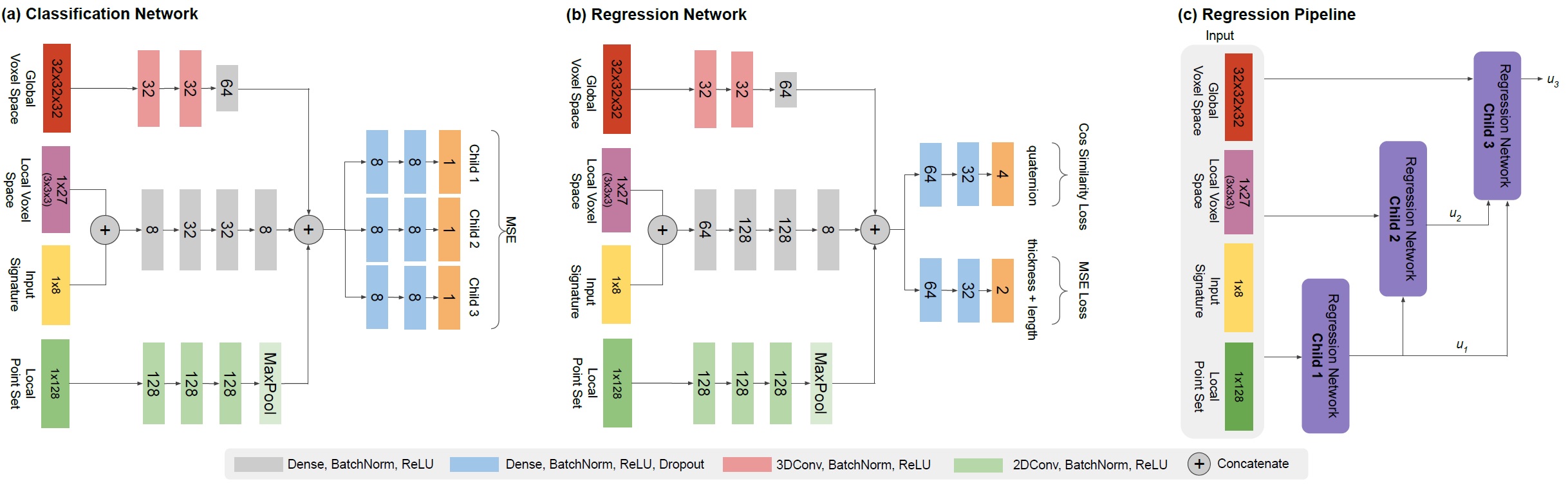}
\vspace{-5mm}
\caption{Network architecture: our neural network pipeline consists of a classification network to predict 0-3 children for a given node (a). Depending on the number of children, we use a cascade of regression networks (c) to predict output signatures for each child. Each regression network is defined as an MLP (b). For each network, we use the current node's signature, the local voxel space for the node, and the global voxel space as inputs. \chh{We add a point encoder network to obtain a feature from local point sets and to enable predicting branchlets for reconstruction tasks.}}
\vspace{-3mm}
\label{fig:network_architecuture}
\end{figure*}

\section{Overview}

Our goal is to learn all possible combinations of branching patterns as they occur from the perspective of a single node in a branch graph. Instead of defining these patterns manually based on parameters and rules, we employ a set of neural networks that learn to predict the child nodes based on the developmental (growth stage, trunk vs. twig, etc.) and environmental context (obstacles, light, gravity, etc.) of a parent node in the branch graph. We train the networks on \textit{branchlets}, a tree node, and its immediate children (Fig.\ref{fig:overview}, a). A~node is defined by all its attributes that we refer to as \textit{node signature}. The node signature serves as input to our network pipeline, and the output is the signatures of up to three child nodes that are then used for the generation of the next nodes. 

The \name\ neural network pipeline consists of a classification network that predicts the topology of a \textit{branchlet}, i.e., how many children a node needs to have. We train individual regression networks for each child node that predict the attribute values of the child node's signature, i.e., their thickness, length, and growth direction. The input of the first regression network is the signature of the active parent node, and it predicts the signature of the first child. We then query the second regression network to predict the signature of the second child. This network receives the signature of the active parent node and the predicted signature of the first child from the previous step as input. This process is continued for the number of children the classifier predicted.

The key idea behind our method is that the networks learn to predict meaningful attributes of child nodes only based on where the parent node -- along with all its signature attributes -- is located within the tree model. Training the network with branchlets enables the network to learn a latent space that encodes which topological and geometric configurations are reasonable for certain locations in the tree's growth space. As an example, for the illustrated tree graph (Fig.~\ref{fig:overview}, a), the networks first need to predict the trunk -- each node only has one child -- for the first few iterations. The further the generation of a tree model progresses, the more the networks need to predict more diverse and finer branching patterns up to the point where only small twigs are generated (Fig.~\ref{fig:overview}, b-d). Eventually, the network also has to predict the termination of branches, which stops the generation of a tree model (Fig.~\ref{fig:overview}, d). 

The objective of our pipeline is not to reconstruct a given tree model. Instead, our goal is to propose an alternative method for defining the rules and parameters commonly used for generating branch graphs with procedural models. A procedural model generates a distribution of tree models based on value ranges defined for the various parameters. A parameter value is obtained at each growth step by sampling from the range. For complex procedural models, the ranges of parameter values can also vary over time. For example, to capture the impact of gravitropism, branching angles may vary from a more narrow parameter range for the trunk to a wider one for the thinner branches in the tree crown. Our method aims to learn these distributions of branching patterns and how they vary across a tree model. We use neural networks to imitate the developmental process locally. 

%% file: src/04-method.tex
\section{Deep Modeling of Branching Structures}

Here we introduce our framework, including a formal definition of the used representation of trees, node signatures, the used neural networks, and the implementation of the environmental response of tree models.   

\begin{figure}[t]
\centering
\includegraphics[width=\linewidth]{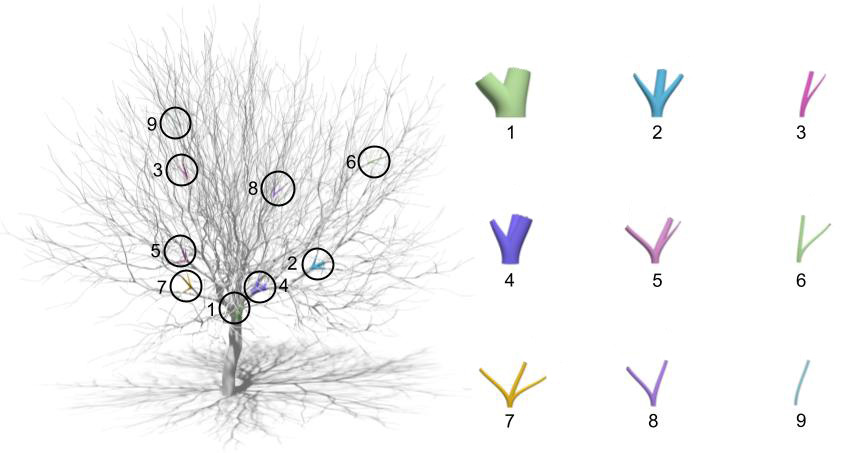}
\vspace{-4mm}
\caption{Illustration of our tree representation: a branchlet is defined as a node with its immediate successors (0-3 children). Branchlets can \ch{have} different geometries, depending on where they are located within a tree model.}
\vspace{-3mm}
\label{fig:branchlets}
\end{figure}

\subsection{Tree Representation}

We represent tree models as a directed acyclic graph $G=\{N,E\}$, where $N$ are the nodes and $E$ the edges. Given an oriented edge $e_i=(n_s,n_t)$ with a starting node $n_s$ and the end node $n_t$, a hierarchical relationship is generated. The node with no predecessor is the root node $n_{root}$. Branches are defined as chains of a varying number of edges $C~=~\{e_1,e_2,\dots, e_m\}$; the length of a chain is denoted as~$m$. We assign to each node the Hack~\shortcite{Hack1957studies} ordering number $o$ \chh{(also called Gravellius ordering)} that assigns $o=0$ to the trunk and one order higher for each branch \chh{recursively (hierarchical ordering)}. Formally, a node is defined as a tuple of attributes 
\begin{equation}
    n = \left(p, t, d_r, d_s, q_p, l_p, o, m_p, a, v, s\right),
    \label{eq:tuple}
\end{equation}
where $p$ represents the 3D position of a node in Euclidean space, $t$~is the branch thickness, $d_r$ is the distance of a node to the root node \chh{in the node hierarchy}, $d_s$~is the distance to the start of the branch (chain), $q_p$ is the normalized quaternion representing the orientation of the node w.r.t. the parent node, $l_p$ is the length of the \chh{edge} to the parent node, $m_p$ is the number of children of the parent node. Additionally, we add the global attributes: age~$a$, gravitropism~$v$, and species~$s$, where $a$ and $s$ are integers. We refer to the tuple $n$ as the \textit{input signature} of a node, which is used as the input of our neural networks. 

A branchlet is defined as a set of edges of a node and its children $B = \{ (n_i, n_{i+1}),  ..., (n_i, n_{i+c})\}$. The number of children of a node is denoted~$c=1,\dots,3$. A branchlet can also have no children  $B=\{ (n_i, \emptyset ) \}$, which is required for terminal nodes. In Fig.~\ref{fig:branchlets}, we show renderings of branchlets with different topological and geometric properties and where they are located within a tree model.

\subsection{Neural Network Pipeline}

Most procedural modeling algorithms aim to produce a graph $G$ to define the properties of a given tree species, which is accomplished by defining the parameters and rules for generating node topology and geometry. For example, one may define the angle at which a branch grows away from its parent branch as a range of values that are then sampled randomly when the procedural model is evaluated to generate a tree model. 

\paragraph{Learning Topology:} Contrary to common procedural approaches, our goal is to learn the generation of \textit{branchlets} -- the branching pattern for a single node $n \in N$ of a branch graph $G$. To successfully generate local branching patterns, we need to simultaneously predict the number of children (topology) and their geometric attributes (e.g., position, thickness, and branching angle). Jointly learning the topological and geometrical properties of a branch graph is challenging as most deep network architectures do not support generating outputs of arbitrary lengths, and the regression of high-dimensional outputs is difficult. Therefore, we learn the number of children and their geometric features separately. To predict the topology of a branchlet, we train a classification network to predict the number of children a given parent node should have as probabilities. More specifically, we aim to learn the mapping:
\begin{equation}
    f_{class}(n): {N} \rightarrow \mathcal{X},
\end{equation}
where ${x} \in \mathcal{X}$ denotes a tuple of probabilities $x=(p_1, p_2, p_3)$ for each child node of $n \in N$. The probabilities determine whether or not to grow a child node. The goal of this network is \ch{only to} predict the topology of a branchlet. 

\paragraph{Learning Geometry:} To learn the geometric properties of a child node, we define the tuple $u = (q, t, l)$, where $q$ denotes a normalized quaternion representing the direction of a child node w.r.t. the parent node, $t$ is the branch thickness of a node, and $l$ the length of the edge between the parent node and the child node. We refer to the tuple~$u$ as the \textit{output signature} of a node and use it as the label for training the regression networks.  

To generate the children of a parent node, we need to predict the geometric properties of up to three tuples $u_1, u_2, u_3$ for each potential child node. We use a cascade of up to three regression networks to predict these tuples. The first network receives only the \textit{input signature} of a parent node as the input and is defined as: 
\begin{equation}
    f_{reg}^1(n): {N} \rightarrow \mathcal{U}_1,
\end{equation}
where $u_1 \in \mathcal{U}_1$ is the output tuple of the first child for a given node $n$. For the subsequent children, the second and third network also receive the predicted output signatures of the previous steps. Consequently, the networks for the second and third child can be defined as: 
\begin{eqnarray}
    f_{reg}^2(n, u_1)&: {N} \rightarrow \mathcal{U}_2,\\
    f_{reg}^3(n, u_1, u_2)&: {N} \rightarrow \mathcal{U}_3,
\end{eqnarray}
where $u_2 \in \mathcal{U}_2$ denotes the output tuple of the second child, and $u_3 \in \mathcal{U}_3$ is the output tuple of the third child for node $n$ respectively. 
Please note that while we did not experiment with predicting more than three child nodes for each parent node, the architecture can be extended as outlined here.

\begin{algorithm}[t]
\SetAlgoLined
\LinesNumbered
\KwIn{Root node $n_{root}$}
\KwOut{Tree graph $G$.}
\textbf{Procedure:}\\
\,\,\,Add $n_{root}$ to $G$.\\
\,\,\,Add $n_{root}$ to list of active nodes $L$.\\
\,\,\,\textbf{while} $|L| > 0$ \textbf{do}\\
\,\,\,\,\,|\,\,\, $n_{a}$ $\leftarrow$  Pop node from L.\\ 
\,\,\,\,\,|\,\,\, ${x}$ $\leftarrow$ Predict probability of each child $f_{class}(n_a)$. \\ 
\chh{
\,\,\,\,\,|\,\, $u_{1} \leftarrow$ Predict output signature of child $1$ \\
\,\,\,\,\,|\,\,\,\,\,\,\,\,\,\,with $f_{reg}^1(n_a)$.\\
\,\,\,\,\,|\,\, $u_{2} \leftarrow$ Predict output signature of child $2$ \\
\,\,\,\,\,|\,\,\,\,\,\,\,\,\,\,with $f_{reg}^2(n_a, u_1)$.\\
\,\,\,\,\,|\,\, $u_{3} \leftarrow$ Predict output signature of child $3$ \\
\,\,\,\,\,|\,\,\,\,\,\,\,\,\,\,with $f_{reg}^3(n_a, u_1, u_2)$.\\
}
\,\,\,\,\,|\,\,\, \textbf{for} $p_i \in {x}$ \textbf{do}:\\
\,\,\,\,\,\,\,\,\,\,|\,\, $\alpha \leftarrow$ Generate a random number in the range $[0, 1]$.\\
\,\,\,\,\,\,\,\,\,\,|\,\, \textbf{if} $\alpha > p_i$:\\
\,\,\,\,\,\,\,\,\,\,\,\,\,\,\,|\,\, Make node $n_i$ \chh{with $u_i$} and add to $G$.\\
\,\,\,\,\,\,\,\,\,\,\,\,\,\,\,|\,\, Push $n_i$ to $L$.\\
\,\,\,\,\,\,\,\,\,\,|\,\, \textbf{end}\\
\,\,\,\,\,|\,\,\, \textbf{end}\\
\,\,\,\textbf{end}\\
\caption{Geometry Generation.}
\label{alg:implement}
\end{algorithm}

\begin{figure*}[t]
\centering
\vspace{-2mm}
\includegraphics[width=\linewidth]{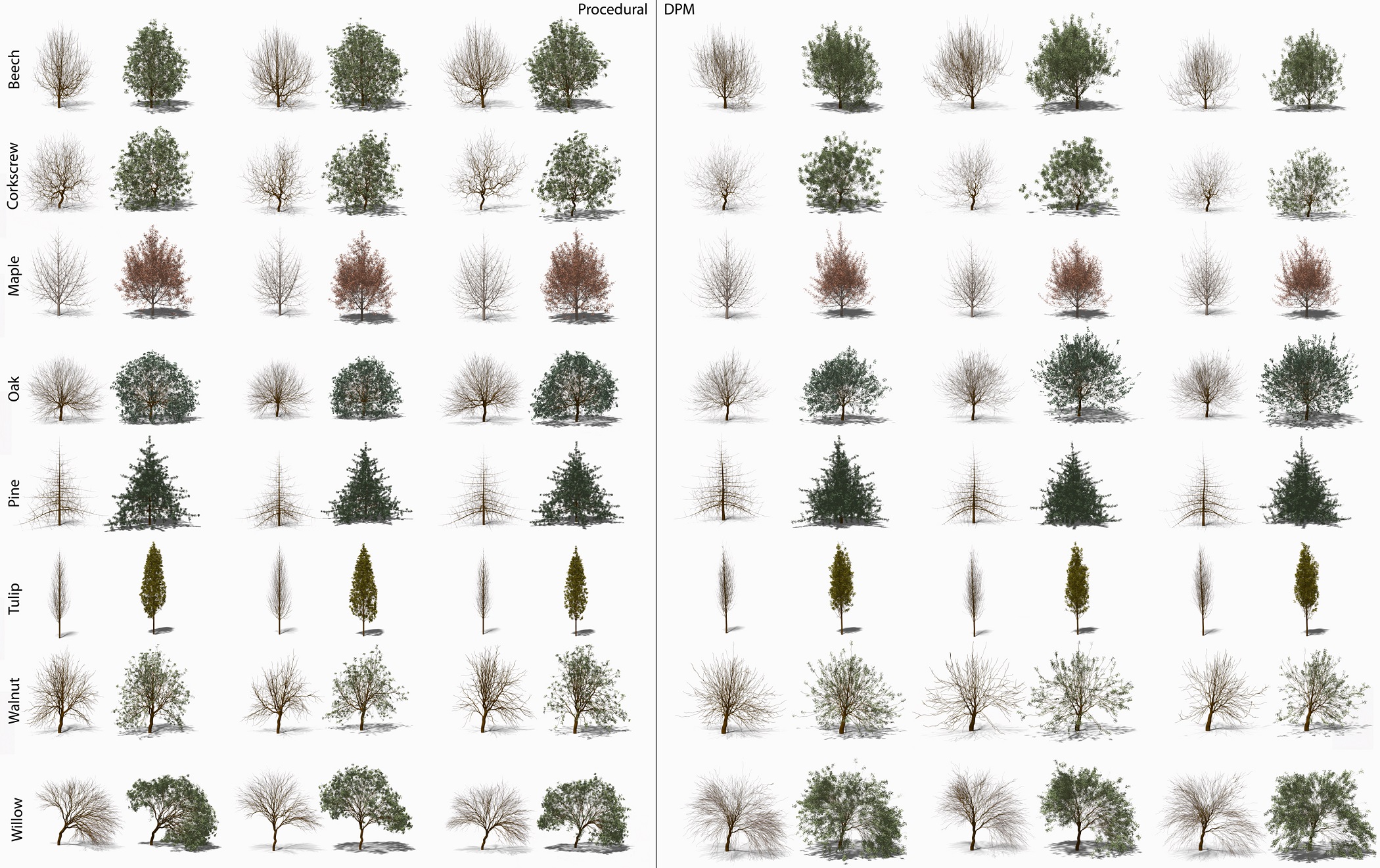}
\vspace{-4mm}
\caption{Eight different tree species modeled with the procedural model of Stava et al.~\shortcite{Stava2014} (left) and with our \name~approach (right). The learned situated latent spaces allow us to generate the topological and geometric details to model a wide variety of different tree species.    
}
\label{fig:species_all}
\end{figure*}

\subsection{Iterative DeepTree Generation}
We iteratively apply our network for the nodes of the branch graph of a tree model to  generate the 3D tree structure. Given the signature of a node~$n$, the classification network predicts the number of children that need to be generated for this node. We then use the cascade of regression networks to predict up to three child nodes. The output of each of the regression networks is an \textit{output signature} containing the quaternion~$q$, the thickness~$t$, and the length~$l$, for each child node of~$n$. Given the position $p$ of a node, we can compute the position of each child node as 
\begin{equation}
    p_{child} = p + l\cdot d_{front}.
\end{equation}
The growth direction $d_{front}$ and the up direction $d_{up}$ are:
\begin{eqnarray}
    \label{eq:dirs1}
    d_{front} &= \boldsymbol{R} \times d_{init\_front}, \nonumber\\
    \label{eq:dirs2}
    d_{up}    &= \boldsymbol{R} \times d_{init\_up},
\end{eqnarray}
where $R$ is the rotation matrix computed from the quaternion $q$. The vectors of the initial directions $d_{init\_front}$ and  $d_{init\_up}$ are $[0, 0, -1]$ and $[0, 1, 0]$ respectively. The vectors $d_{front}$ and $d_{up}$ are used to generate the branch mesh. The thickness is stored with the child node and later also used to generate the branch mesh. 

To generate a new tree model, a user defines the signature of the node $n_{root}$. The root node is added to a list of active nodes~$L$. As long as the list holds nodes, an active node $n_a$ is popped from the list. 
The classification network then predicts which of the three children should be generated; the result is returned as the tuple \chh{$x = (p_1, p_2, p_3)$}. If a child node needs to be generated for $n_{a}$, the cascade of regression networks predicts the tuple $u_i$ for each child. The child nodes are added to the branch graph, and the list of active nodes and the network pipeline can be queried for each node in the subsequent iterations. The generation of a branch terminates when the classification network predicts zero children for a node. Pseudocode for our algorithm is shown in Alg.~\ref{alg:implement} and an illustration of the iterative construction of a tree model is shown in Fig.~\ref{fig:overview}.

\begin{figure*}[t]
\centering
\includegraphics[width=\linewidth]{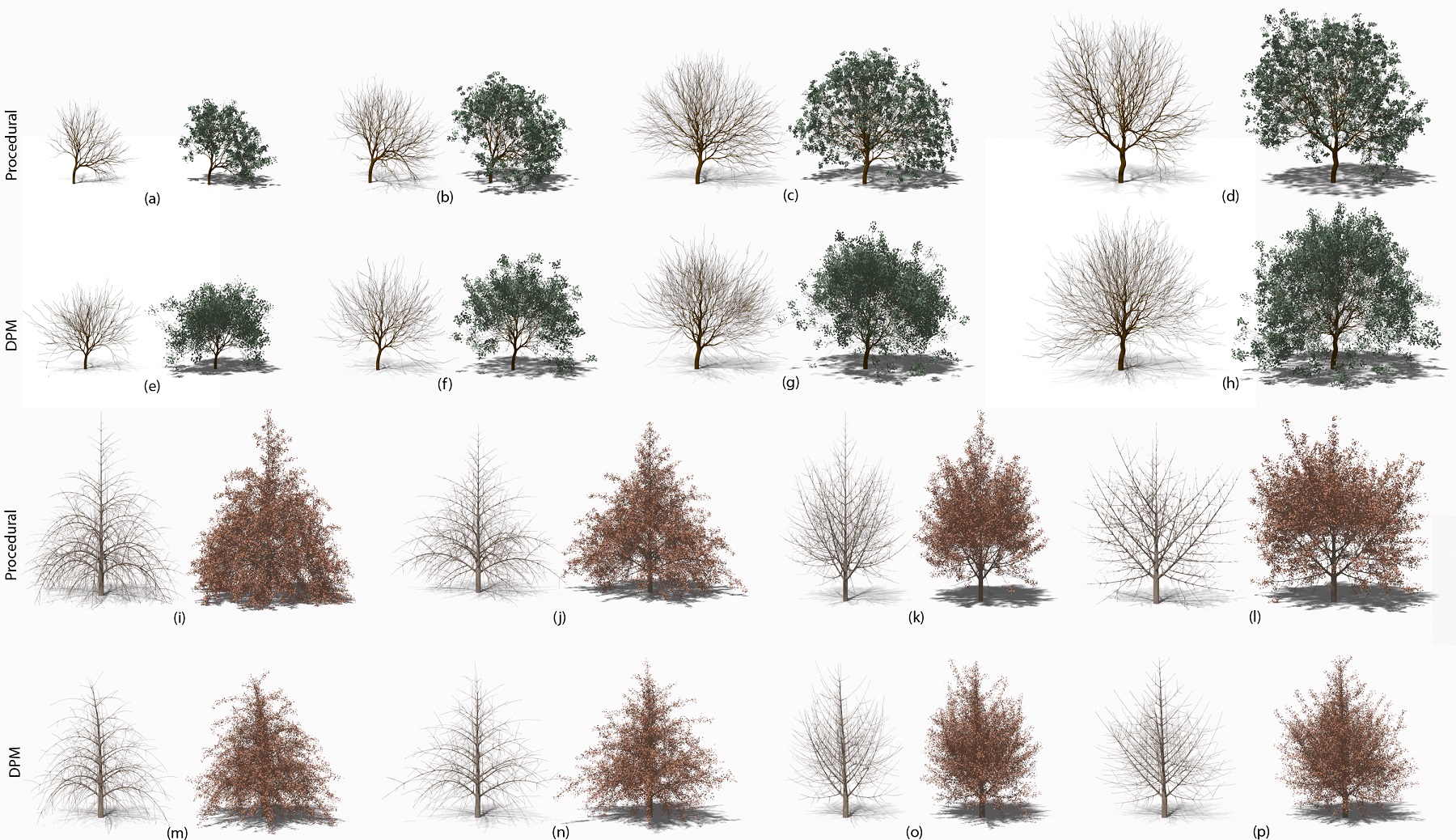}
\vspace{-3mm}
\caption{
Modeling age and gravitropism: by adding priors for age and gravitropism to our node tuple $n$, our neural network pipeline is able to model realistic branching structures. Here we compare a procedurally generated Oak tree at different developmental stages (a-d) and the corresponding \name~generated ones (e-h). Our method can also capture the differences in branching patterns when considering different environmental factors, such as gravitropism. The procedurally generated tree models (i-l) show the same structural properties as the \name~generated ones~(m-p). 
}
\label{fig:prior_age_gravity}
\end{figure*}

\subsection{Environmental Response}

We use a voxel space to learn how trees grow around obstacles in their environment~\cite{Greene:89:SIGG}. We represent obstacles as bounding boxes and generate an occupancy grid $Y$ that encodes where obstacles are located in the environment. We generate the occupancy grid by testing the bounding boxes against the voxels. We use the computed occupancy grids in two different ways. First, we obtain the $3\times{3}\times{3}$ neighborhood of cells for each node as 27 binary occupancy values denoted as~$Z$. We then add them to our node signature $n = n \cup \{Z\}$. As shown in Fig.~\ref{fig:network_architecuture} (a-b), we can use these values as additional input for our classification and regression networks (magenta boxes). Second, to obtain a global feature of the environment, we use an additional encoder consisting of three 3D convolutional layers. \ch{The global voxel space is a one-hot 3D vector where obstacle voxels are marked as one and empty voxels as zero; } \chh{it has a size of $32^3$ voxels.} \ch{The tree's root is set in the middle of the voxel space.} We extract the global feature through the additional encoder and added it to the local feature before the dense layers for the output heads of the networks. Adding the global feature of the environment to our neural networks allows us to generate environmentally sensitive tree models that can be placed arbitrarily into the environment. In contrast to more complex models of environmental sensitivity~\cite{Palubicki:2009:STM,Mech:96:SIGG,Pirk:2012:PTI:2185520.2185546}, this approach provides a lightweight way to generate realistic branching structures for tree models placed in the vicinity of obstacles.

\subsection{\chh{Point-based Modeling}}
\label{label:point_modeling}
\begin{wrapfigure}{r}{0.5\linewidth}
\begin{centering}
\vspace{-3mm}
\hspace{-5mm}
  \includegraphics[width=\linewidth]{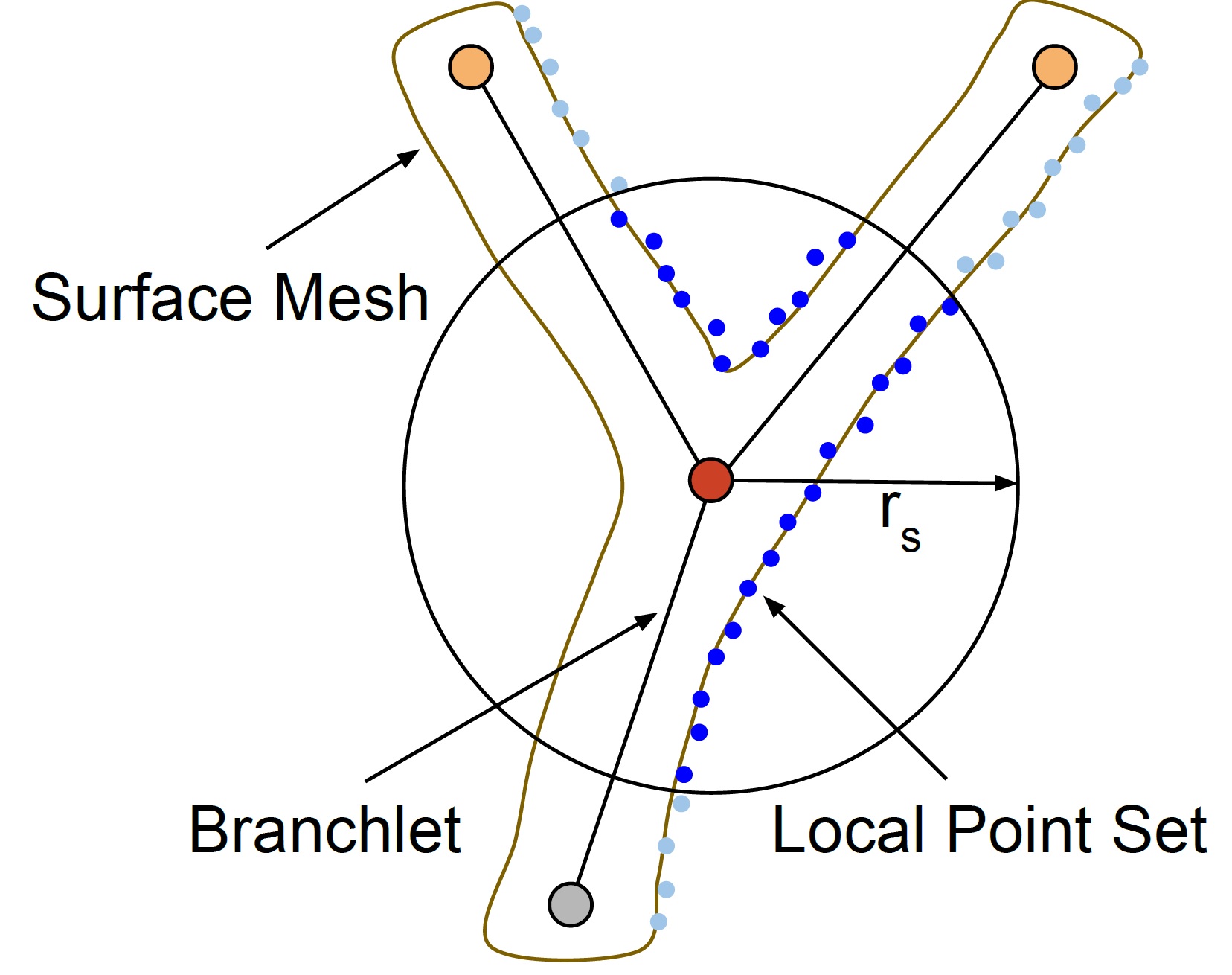}    
  \label{fig:branch_points}
\end{centering}
\vspace{-3mm}
\end{wrapfigure}
To provide additional control for the reconstruction and sketch-based generation of tree models, we have equipped our neural network architecture with an additional point-cloud encoder network 
\chh{(Fig.~\ref{fig:network_architecuture} a-b). We locally sample points from a point cloud of a procedurally generated tree model to train our architecture for tree reconstruction with point clouds. For each node of a branchlet, we define a sample radius $r_s$ to compute a sphere around a node (inset figure). Given a point cloud of a tree, we select all points within the sphere; the obtained local point set is denoted as $P$. We then associate the local point set to our input signature $n = n \cup \{P\}$ and use it as input to our architecture to obtain a point feature. The point feature is then added to the signature embedding for the classification and regression networks. This training setup aims to condition the prediction of child node attributes based on the captured local point sets. The implementation of our point encoder follows PointNet~\cite{8099499}. For most of our experiments, a radius of $r_s=2.0$ provides the best results for reconstructing the tree models.}

%% file: src/05-results.tex
\section{Implementation}

Our framework is implemented as two components: the first is an interactive framework to efficiently generate large collections of tree models with the procedural model from~\cite{Stava2014}. Second, we use Pytorch for developing and training our classification and regression neural networks. Below we describe how we used these two components of our framework to generate training data and how we trained our neural network pipeline. 

\chh{The deep learning model was trained on a single Nvidia RTX A5000, and it took 6 hours for the regressor and 4 hours for the classifier. Our rendering was performed on a desktop computer equipped with an Intel(R) Core(TM) i9-9900K and Nvidia 3090 RTX.}

\begin{figure*}[t]
\centering
\includegraphics[width=\linewidth]{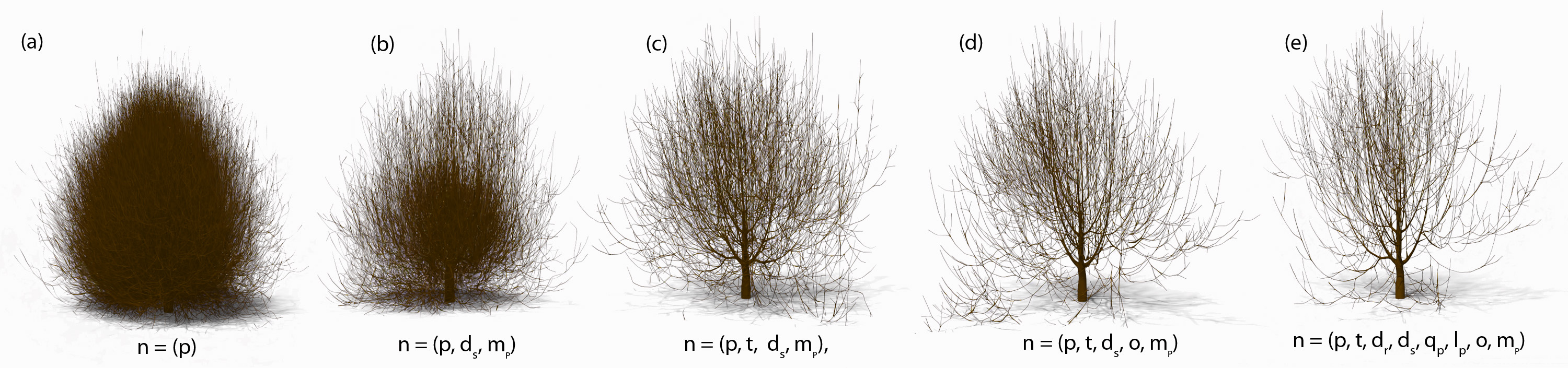}
\vspace{-7mm}
\caption{Ablation study of the classification network: here, we show the impact of different attribute configurations of our node signature as an ablation study. Training only with the attribute position ($p$) generates a dense branching structure, and the network fails to terminate (a). Even if additional attributes such as the root distance ($d_s$) and the number of children of the parent node ($m_p$) are added, the model still fails to terminate (b). c) and d): only after adding additional attributes, such as the thickness ($t$) or branch order ($o$), the model generates more convincing branching structures, although some branches are still generated into the ground. 
Please note that we only use the per-node attributes for this experiment and not the set of global attributes such as species ($s$), age ($a$), and gravitropism ($v$). \chh{The tree species we used for this experiment is Beech (Fig.\ref{fig:species_all}).}
}
\vspace{-2mm}
\label{fig:ablation}
\end{figure*}
 \begin{figure}[t]
\centering
\includegraphics[width=\linewidth]{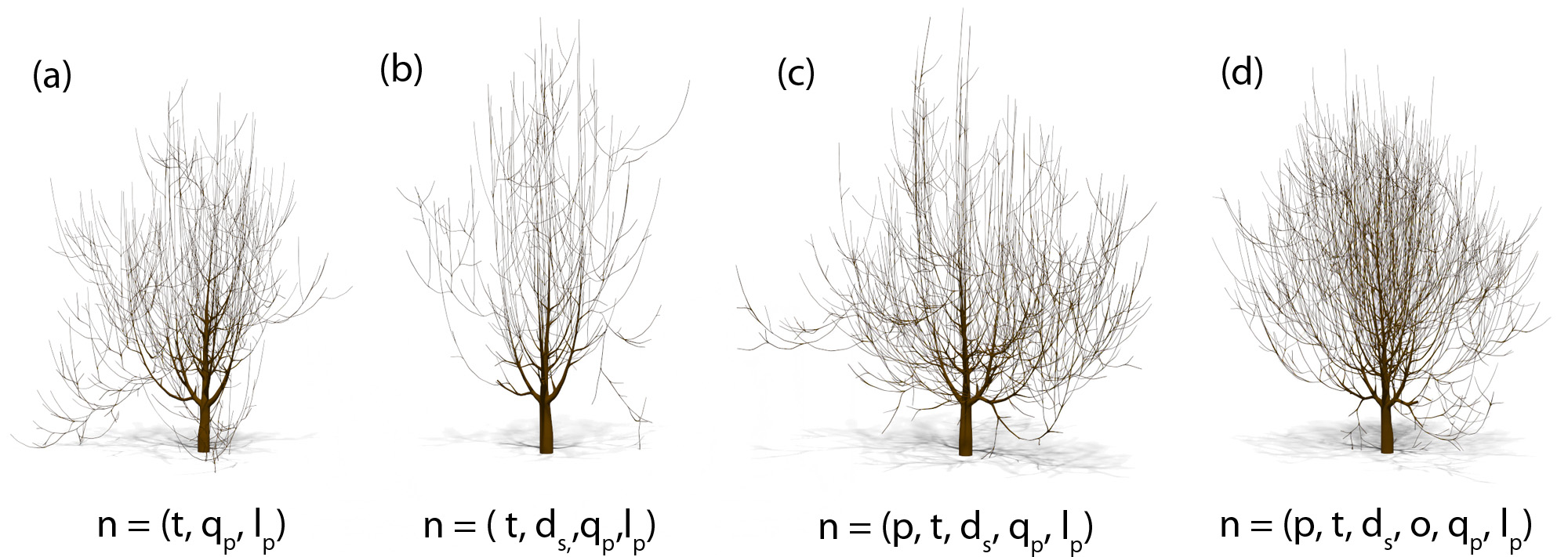}
\caption{Ablation study of regression networks: the selection of node attributes for training the regression networks can profoundly impact the generated branch graphs. Only using a few attributes leads to slim and sparse branching structures (a, b). Adding more attributes to the input signature, such as the position or the order (c, d), will lead to more realistic branching patterns. For this experiment, we used a classification network with all parameters. The model with all node attributes is shown in Fig.\ref{fig:ablation} (e).}
\label{fig:ablation_reg}
\end{figure}

\subsection{Dataset Preparation}
\label{sec:dataset_training}

We used the 29-dimensional parameter space of the procedural model from \citeauthor{Stava2014}~\shortcite{Stava2014} to model eight distinct tree species: Beech, Corkscrew, Maple, Oak, Pine, Tulip, Walnut, and Willow (Fig.~\ref{fig:species_all}, left). An explanation of the parameters of the procedural model, along with the used parameter values, can be found in the Appx. Tab.~\ref{tab:parameters} and Tab.~\ref{tab:parameter_values}. We then generate 500 unique tree models for each species and normalize their positions (attribute $p$) into a unit cube. We split these tree models into 400 for training and 100 for validation. \chh{The validation data is used for identifying optimal hyperparameters for our neural network architectures (e.g., learning rate, number layers, etc.).} The procedural model and our \name~algorithm only generate branch graphs. We generate a mesh out of generalized cylinders based on the branch graph and the stored thickness values to render images of trees with high visual quality. The voxel space that we use for representing the environment has $32\times{32}\times{32}$ voxels. We compute the bounding box of each obstacle in the environment (e.g., walls) and check whether they occupy the corresponding voxels. This results in a 3D occupancy grid.   

\chh{To train the point cloud network, we generate view-dependent (partial) point clouds for our procedurally generated tree meshes. To generate the point scans, we mimic a LiDAR scanner that we place on a 10m$\times$10m ground plane outside the bounding box of a tree that is located at the origin of the plane. We then cast rays toward the tree by sampling points on a sphere located at the scanner position. Each ray is tested against the surface mesh of a procedural tree model of our dataset. This procedure generates partial point scans of our tree models with 8k-35k points. We then sample this point cloud for each branchlet to obtain a local point cloud with a maximum of 512 points as described in Sec.~\ref{label:point_modeling}}. 

 \subsection{Training}
We train the classification and regression neural networks of our pipeline based on \textit{branchlets}. We generate the branchlets by decomposing the generated branch graph $G$. \chh{Any node in the tree graph (with 0-3 children) can serve as a branchlet. We traverse each graph in our training dataset and select each node along with its immediate children to generate a set of branchlets $\mathcal{B}$.}
A parent node of a branchlet with all its attributes becomes the \textit{input signature}. We then compute the quaternion $q$ and the length $l$ based on how child nodes are connected to the parent node of the branchlet and store these attributes along with the thickness value $t$ as the \textit{output signature} that is used as a label for training the regression networks. For training the classification network, we simply obtain the number of children attached to the parent node as the label. 

\textit{Classification Network Training:} we train our classification network for 150 epochs jointly on the 3,200 tree models of all species. As illustrated in Fig.~\ref{fig:network_architecuture}, our classification network has three output heads. The last layer uses a sigmoid activation function. For each head, we aim to simultaneously predict the probability of whether a particular child needs to be generated (0=child is omitted, 1=child is added). We train for this objective with an MSE loss, where the label for adding a child is $1$ and $0$ to omit a child node. \chh{Depending on the species the network is trained on we obtain a child classification accuracy of 83 - 93\%}. 

\begin{figure*}[t]
\centering
\vspace{-1mm}
\includegraphics[width=\linewidth]{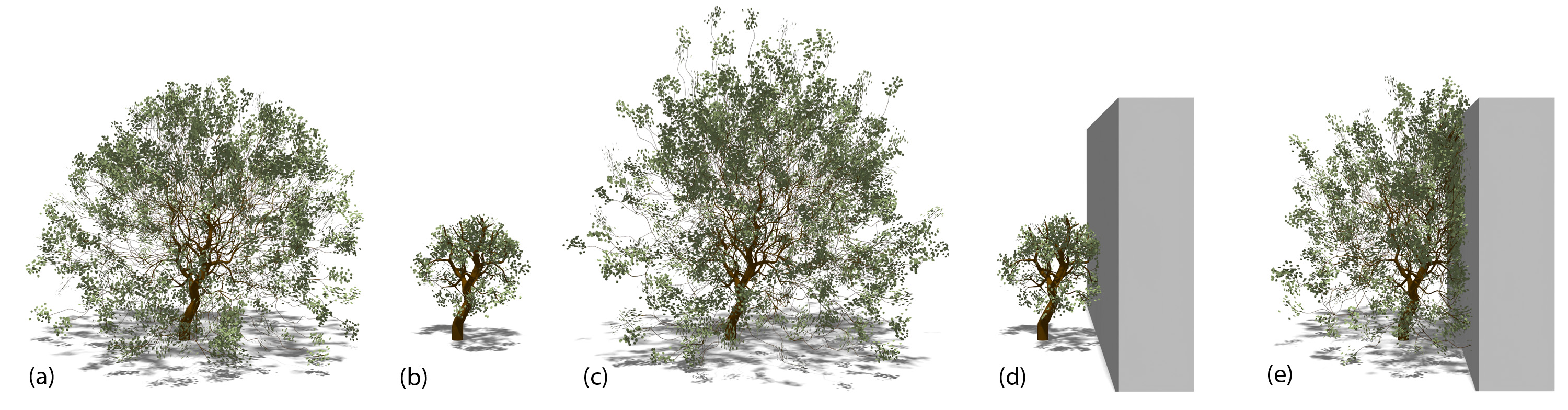}
\vspace{-9mm}
\caption{Regrowth: our method can be used to regrow tree models to thereby adapt them to changing environmental conditions. A fully developed tree model (a) generated with our method is pruned and can be developed anew (b) to a fully grown tree model (c) by iteratively applying our \name~pipeline to the outermost nodes in the branch graph. If the tree model is placed in the vicinity of an obstacle (d) and then regrown (e), our method supports generating \ch{environmentally sensitive tree models.}}
\vspace{0mm}
\label{fig:regrowth}
\centering
\includegraphics[width=\linewidth]{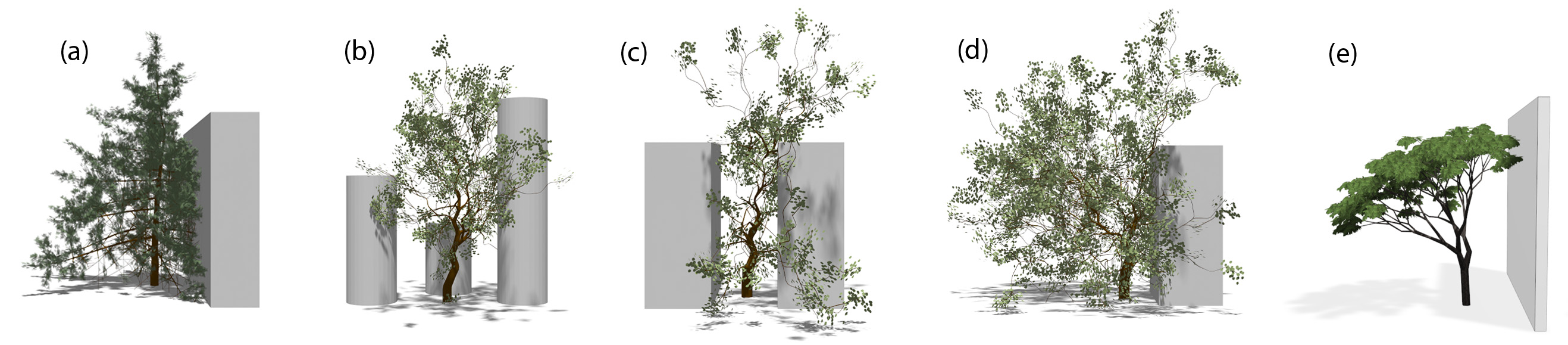}
\vspace{-8mm}
\caption{Modeling with environmental sensitivity: when provided with the environment as occupancy grids, our situated latent space is able to generate convincing branching structures to mimic the growth response of trees to obstacles in their vicinity (a-d). \chh{Similar to existing approaches, such as \citeauthor{Pirk:2012:PTI:2185520.2185546}~\shortcite{Pirk:2012:PTI:2185520.2185546} (e), our method is also able to generate environmentally-aware branching structures.}}
\vspace{-0mm}
\label{fig:environmental_response}

\centering
\includegraphics[width=\linewidth]{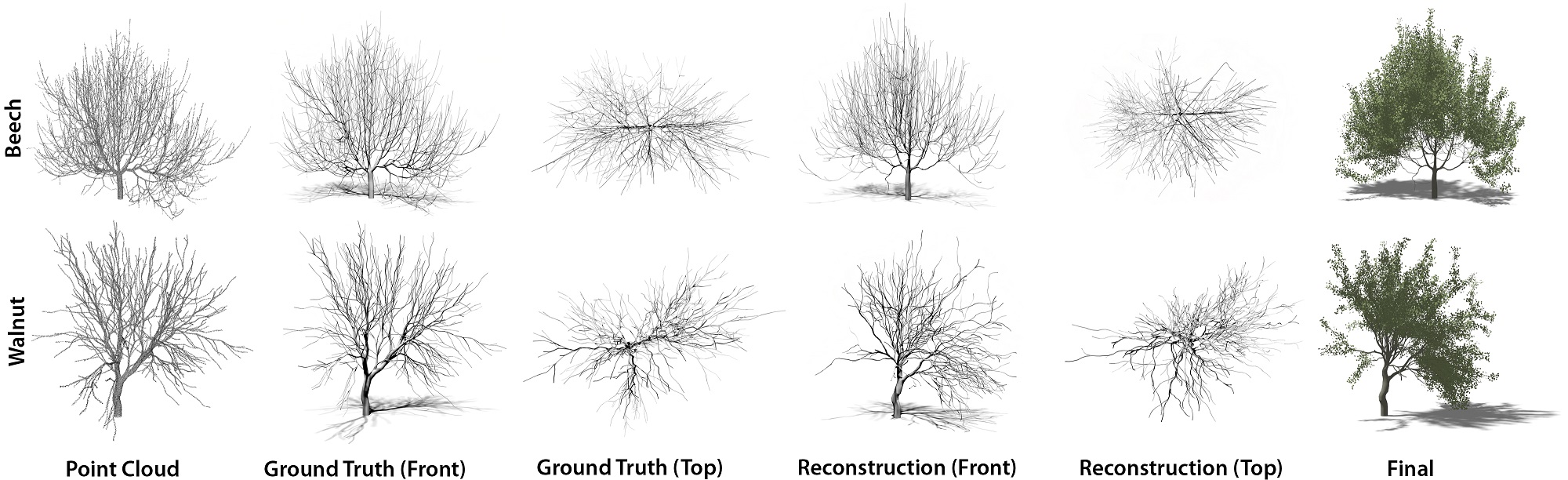}
\vspace{-5mm}
\caption{\chh{Reconstruction of tree models from point cloud data: we sample our procedural trees to generate point clouds. We then train our neural networks with local point clouds and use them to reconstruct unseen tree models from the validation dataset.}}
\vspace{-3mm}
\label{fig:reconstruction_synthetic}

\end{figure*}

To train this network more robustly, we use data augmentation. We add random noise to the position with a mean of zero and std of 0.0001. Additionally, we ensure a balanced sampling of nodes of the same properties. \ch{Considering that nodes with a different number of children appear with different frequencies (e.g., most nodes may have one child), we balance the number of nodes of each type during training to avoid network overfitting.} For example, nodes at the trunk are rarely present in our dataset compared to the many twig nodes in the tree crown. Balancing the nodes helps to avoid overfitting. Training the classification network takes about four hours. 

\textit{Regression Network Training:} each regression network is trained individually for 200-300 epochs on the 3,200 tree models of all species. We trained the regression networks for up to 12 hours. Each regression network has two output heads: the first generates outputs for thickness and length. We train this head with an MSE loss. \chh{The second head predicts a quaternion and is trained with a cosine similarity loss.}
\chh{To avoid overfitting, we also add random noise to the position for training the regression networks with a mean of zero and std of 1e-4. Depending on the species the network is trained on, we obtain an MSE of 1.2e-4 - 3e-4 (branch length and thickness) and a cosine similarity of 0.04 - 0.20 (quaternion).}
The networks are trained with the Adam optimizer, a learning rate of 1e-4, and a batch of 512. An illustration of our network architectures -- also including the used number of units in each layer -- is shown in Fig.~\ref{fig:network_architecuture}. Please note that each of the used regression networks in the cascade of networks (Fig.~\ref{fig:network_architecuture}, c) is trained individually. 

\subsection{Rendering}
We use the generated branch graph from the procedural model and \name~to compute a surface mesh of branches from generalized cylinders. Leaves are represented as textured quads placed along with the outermost branches with additional procedural parameters for \ch{their placement (e.g., phyllotaxis)~\cite{deussen2005digital}}. All models were rendered with a path tracer written with the Nvidia OptiX 7.4 API~\cite{10.1145/1778765.1778803} tracing 512 rays per pixel. 

\section{Results, Evaluation, and Applications}

\chh{Here, we discuss experiments on how our framework can be used to generate tree models of different species, ages, or tropisms. We also discuss applications for our method and show how we validate our framework based on geometric and perceptual metrics.}

\subsection{Results}
Fig.~\ref{fig:species_all} shows a qualitative comparison of eight procedurally generated tree species and the same eight species generated with \name. Our method can capture the wide variety of distinct features across the eight tree species, including the fine nuances in shape and geometric detail. Generating these species with a procedural model is a significant modeling effort that requires manually specifying the procedural model and carefully fine-tuning all the parameters for each species. In contrast, our method learns the branching patterns of all species from data by training our network pipeline. 

The results shown in Fig.~\ref{fig:prior_age_gravity} indicate that our neural network pipeline and the training of situated latent spaces enable capturing the growth response of tree models and the impact of gravitropism. By adding global priors for age $a$ and gravitropism $v$ to the node tuple~$n$, our pipeline can faithfully generate the structural details for modeling different age stages (e-h) and varying degrees of gravitropism (m-p). Compared to the procedurally generated branching structures (a-h) and (j-l), our method can generate branching patterns with almost identical topological and geometric features. 

We conducted ablation studies to analyze the impact of different attributes in our node tuple $n$ during training of the classification and regression networks. Fig.~\ref{fig:ablation} demonstrates the impact of using more or fewer parameters during the training of the classifier. For the result shown in Fig.~\ref{fig:ablation}~(a), we only used the position $p$ attribute for our classification network. Training \chh{the network only on this attribute} prevents the network from terminating, which results in very dense and unrealistic branching patterns. Adding the distance to the branch root $d_s$ and the number of child nodes of the parent node $m_p$ by itself also does not enable training of more realistic branching structures (Fig.~\ref{fig:ablation}~b) as the model still fails to predict the correct number of children. However, by adding the branch thickness~$t$, the network starts to successfully predict the termination of branches, which leads to more realistic, still dense branching structures (Fig.~\ref{fig:ablation}~c). When also using the other attributes, the model starts predicting branchlet topologies and geometries that lead to realistic branching patterns (Fig.~\ref{fig:ablation}~d, e). Please note that for (a) and (b), we manually stopped the iterative generation of the models as the neural networks would not automatically terminate. In Fig.~\ref{fig:ablation_reg}, we show a similar ablation study for the regression networks. For this experiment, we used a classification network with all node attributes (Fig~\ref{fig:ablation}, e).  Only using a few attributes for training the regression networks leads to slim and sparse branching structures (a, b). Adding more attributes generates more realistic branching patterns (c, d).

The result in Fig.~\ref{fig:regrowth} shows the usefulness of our method for content creation. After we generate a fully developed tree model with our method (a), we can prune and regenerate the branch graph (b) to another fully developed tree model (c). If the model is placed next to an obstacle and then regenerated, it adapts to its new environment~(d) and develops an adapted and realistic branch graph~(e). Generating a new tree model by pruning  a branch graph can be repeated indefinitely. However, please note that each generated model is unique -- our method cannot regenerate precisely the same model. 
\begin{figure}[t]
\centering
\includegraphics[width=\linewidth]{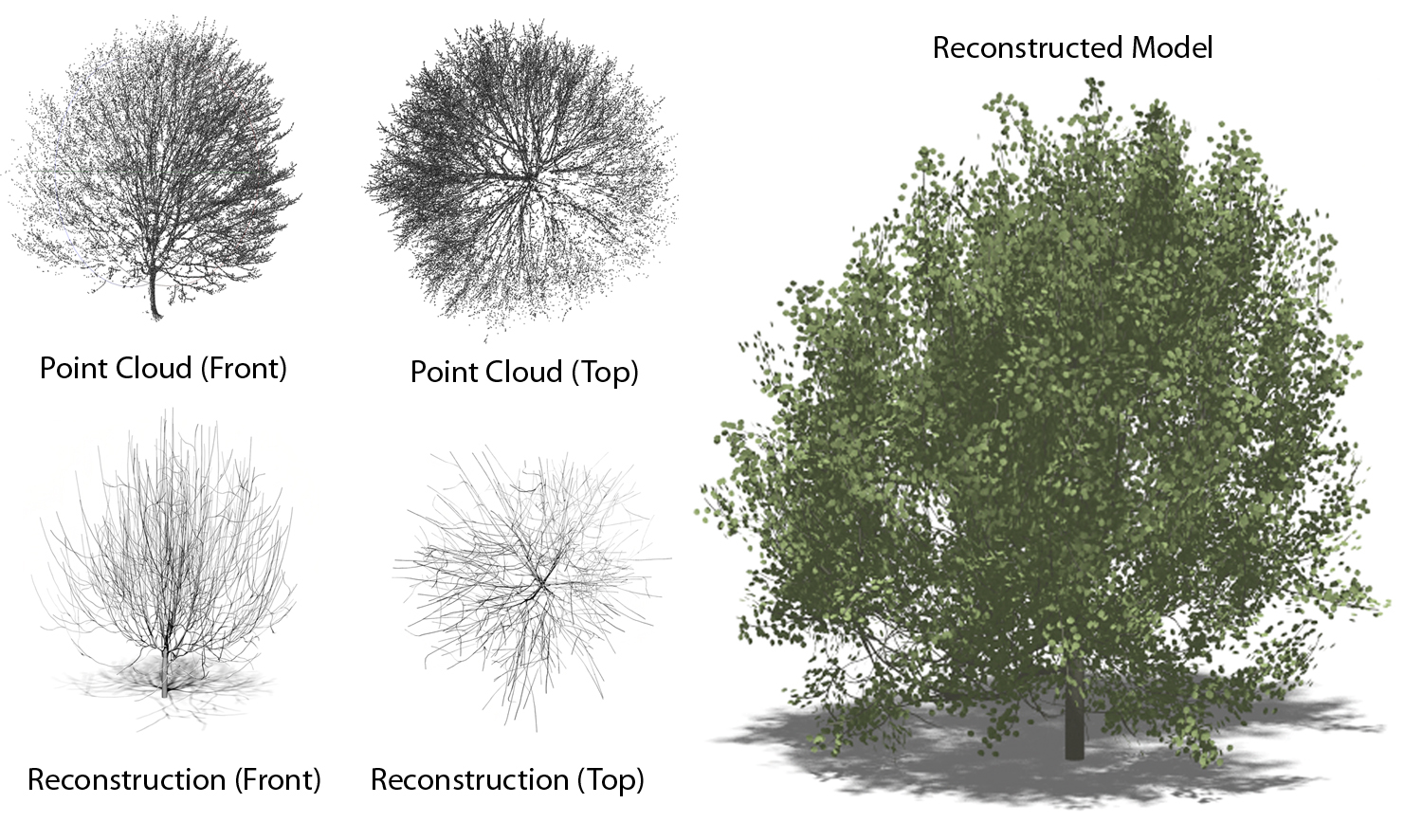}
\vspace{-8mm}
\caption{\chh{Reconstruction of a real tree from captured point cloud data: \name~was trained on synthetic data and generated plausible tree models from real point clouds.}}
\vspace{-4mm}
\label{fig:reconstruction_real}
\end{figure}

\begin{figure}[t]
\centering
\includegraphics[width=\linewidth]{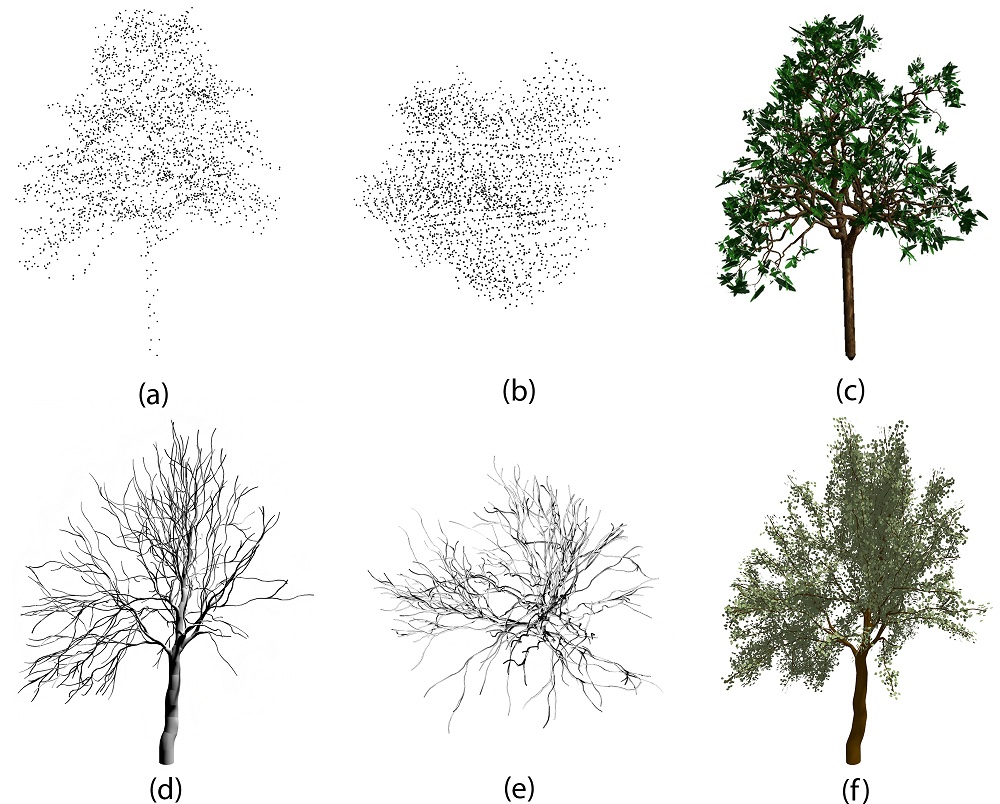}
\vspace{-6mm}
\caption{\chh{Comparison to AdTree~\cite[Fig 11 e]{du2019adtree}: we use an existing point cloud from their paper (a-b) and show their reconstruction (c). \name\ generates a similar tree model (d-f).}}
\vspace{-2mm}
\label{fig:adtree_comparison}
\end{figure}

 \begin{figure}[t]
\centering
\includegraphics[width=\linewidth]{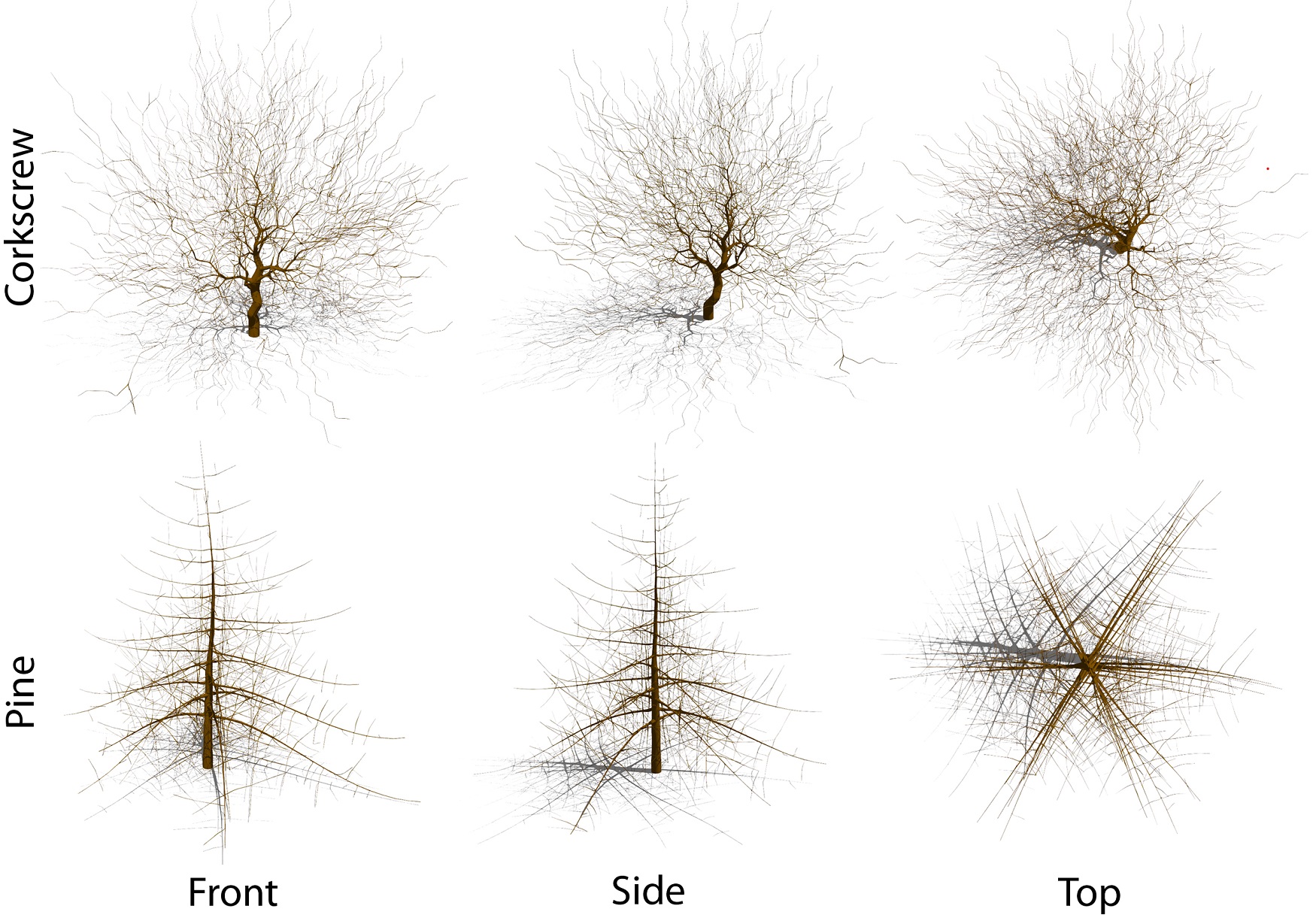}
\vspace{-5mm}
\caption{\chh{Different viewpoints of tree models show that \name~generates tree models without dominant artifacts (e.g., half-reconstructed model) and captures tree phyllotaxis.}}
\vspace{-2mm}
\label{fig:multi_view}
\end{figure}

\begin{table}[t]
\caption{\chh{Tree model compression: our framework can regenerate the same tree model from a root node by setting the same random seed. We show the memory footprint of the root node (Seed Size = 52B) along with the weights required for our neural networks (NN Size) and compare it with the average memory footprint of a single tree model, which can either be stored as a skeletal graph (Graph Size) or surfaces mesh (Mesh Size). We show the average and standard deviation of the size comparison for the graph and mesh size, having the seed and the NN as the base. The graph size is, on average, 1.58$\times$ larger, and the mesh size is about 20.81$\times$.}}
\vspace{-2mm}		
\label{tab:compression}
\scalebox{0.7}{
\begin{tabular}{lccccr}
\toprule
\textbf{Species} & \textbf{Seed + NN Size} & \textbf{Graph Size} & \textbf{\# Nodes} & \textbf{Mesh Size} & \textbf{\# Vertices} \tabularnewline
\midrule
Beech     &  913 kB & 1.85 MB & 5.7k & 24.08 MB & 115k\tabularnewline
Corkscrew &  913 kB & 1.20 MB & 3.7k & 15.66 MB & 75k\tabularnewline
Maple     &  913 kB & 1.36 MB & 4.2k & 17.68 MB & 85k\tabularnewline
Oak       &  913 kB & 0.78 MB & 2.4k & 10.18 MB & 49k\tabularnewline
Pine      &  913 kB & 1.81 MB & 5.5k & 23.86 MB & 114k\tabularnewline
Tulip     &  913 kB & 1.84 MB & 5.6k & 24.27 MB & 115k\tabularnewline
Walnut    &  913 kB & 0.74 MB & 2.3k & 10.97 MB & 53k\tabularnewline
Willow    &  913 kB & 1.60 MB & 4.9k & 20.96 MB & 101k\tabularnewline
\midrule
Size (avg,std)     & [1,0]   & [1.58,0.46] & - & [20.81,5.78] & - \tabularnewline
\bottomrule
\end{tabular}
}
\vspace{-2mm}
\end{table}

Fig.~\ref{fig:environmental_response} shows that our situated latent spaces can encode the environmental response of a tree model. By adding the local and global occupancy features, the neural network pipeline can be trained to model the growth response of tree models to obstacles in their environment. For this experiment, we also generated a dataset of 500 scenes (400 training, 100 validation) where obstacles were randomly placed in the vicinity of the procedurally generated trees. The procedural model accounts for the obstacles by computing the availability of light in its environment -- branches then grow into regions where more light is available. By training with tree models and the occupancy grid from the scene, the neural network pipeline is able to mimic the modeling of the environmental response. Fig.~\ref{fig:environmental_response} (a)-(d) compares our results to the method of \ch{ \citeauthor{Pirk:2012:PTI:2185520.2185546}~\shortcite{Pirk:2012:PTI:2185520.2185546}~(e)}, where environmental sensitivity is modeled based on an inverse algorithm. Our \name~learning-based approach can generate a similar response of adapted branching structures for trees grown in the vicinity of obstacles. 

\subsection{Applications}
\chh{Fig.~\ref{fig:reconstruction_synthetic} shows a tree reconstruction result. We use our training data set to generate point clouds of procedurally generated tree models. We then train our neural network pipeline on local point clouds to provide the network with a point feature (Sec.~\ref{label:point_modeling}). Training with the point feature allows us to generate branching structures that follow the scanned points, which also works for point clouds of real trees, as shown in Figs.~\ref{fig:reconstruction_real} and~\ref{fig:adtree_comparison}}. 

\begin{figure}[t]
\centering
\includegraphics[width=\linewidth]{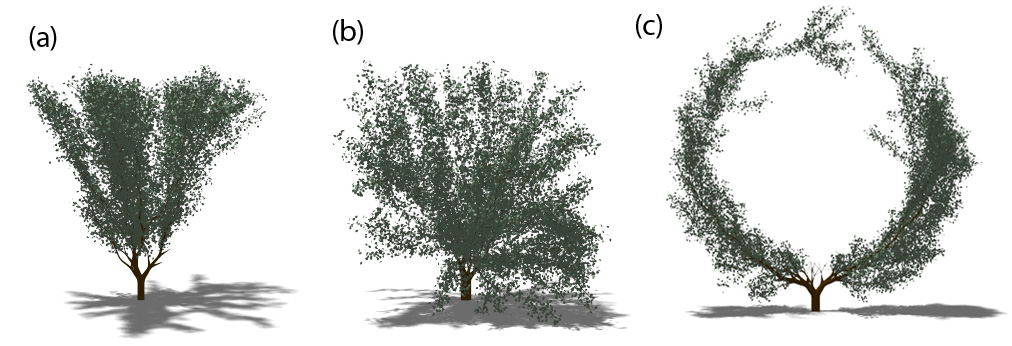}
\vspace{-6mm}
\caption{Our method can also replicate the growth process of form-controlled tree models. For this result, we trained on a dataset of tree models that were grown into meshes of cones (a), cubes (b), and rings (c).}
\vspace{-4mm}
\label{fig:form_control}
\vspace{8mm}
\centering
\includegraphics[width=\linewidth]{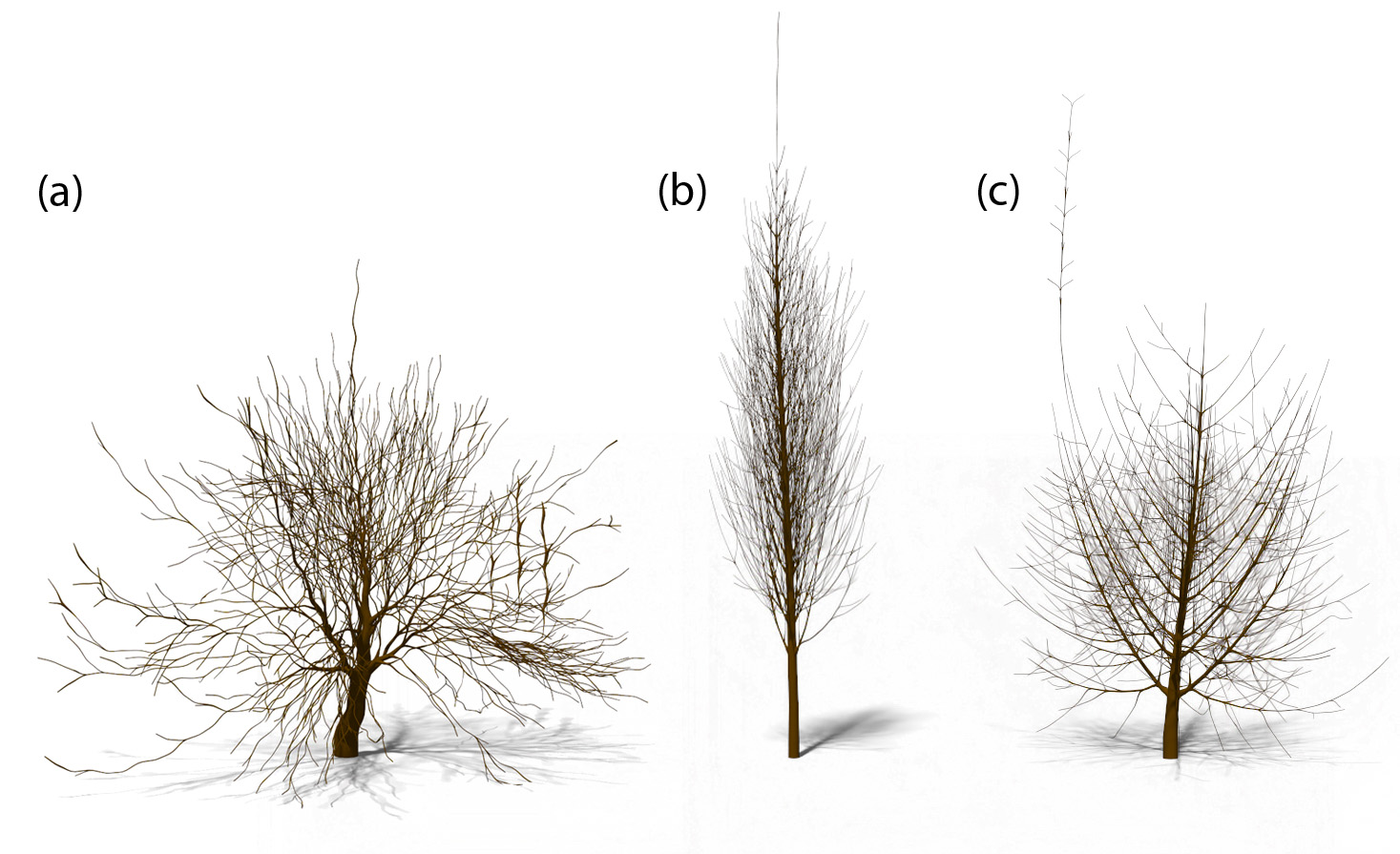}
\vspace{-5mm}
\caption{Three failure cases: if the used neural networks are not properly trained our method tends to \ch{generate branching structures with too thin and too long branches.} Here we show examples of Oak (a), Tulip (b), and Maple (c) trees.}
\vspace{-2mm}
\label{fig:failture_cases}
\end{figure}

\chh{We show multiple views of the same tree models in Fig.~\ref{fig:multi_view} to demonstrate that our method generates plausible tree models from all view directions. Once trained, our framework enables the generation of tree models of a specific tree species. Moreover, \name~generalizes the branchlet and not the entire tree. Thus, each generated model is different. This makes our approach suitable for generating large datasets of new models, as shown in the selection of randomly sampled tree models in Fig.~\ref{fig:stochasticity}.} Fig.~\ref{fig:form_control} shows that our method can replicate the growth of more constrained tree shapes. For this result, we generated a dataset of 500 tree models for each of the shown shapes, including cone (a), cube (b), and ring (c). 

\chh{Finally, in Tab.~\ref{tab:compression} we show that our method can also be used for compressing generated tree models. We compare the memory footprint of a tree model generated by our method to common representations for tree models, such as skeletal graphs and surface meshes. \name~can deterministically re-generate the same tree model by using the same random seed and replacing the stochastic probability with a given number. As we only need to store the root node and the weights for our neural networks, our method allows us to compress tree models with a lightweight memory footprint that is even smaller than most skeletal graphs.}

\subsection{Evaluation}
\ch{We are not aware of any deep neural generative model for 3D tree geometry. Therefore we compare our algorithm to the state-of-the-art procedural model of~\cite{Stava2014,Li2021ToG}.} 
We used 500 procedurally generated tree models (P) for each species, and we generated another 500 trees using \name~(DT). We then compare their geometry and perceived level of realism.

\begin{figure*}[hbt]
\centering
\includegraphics[width=0.99\linewidth]{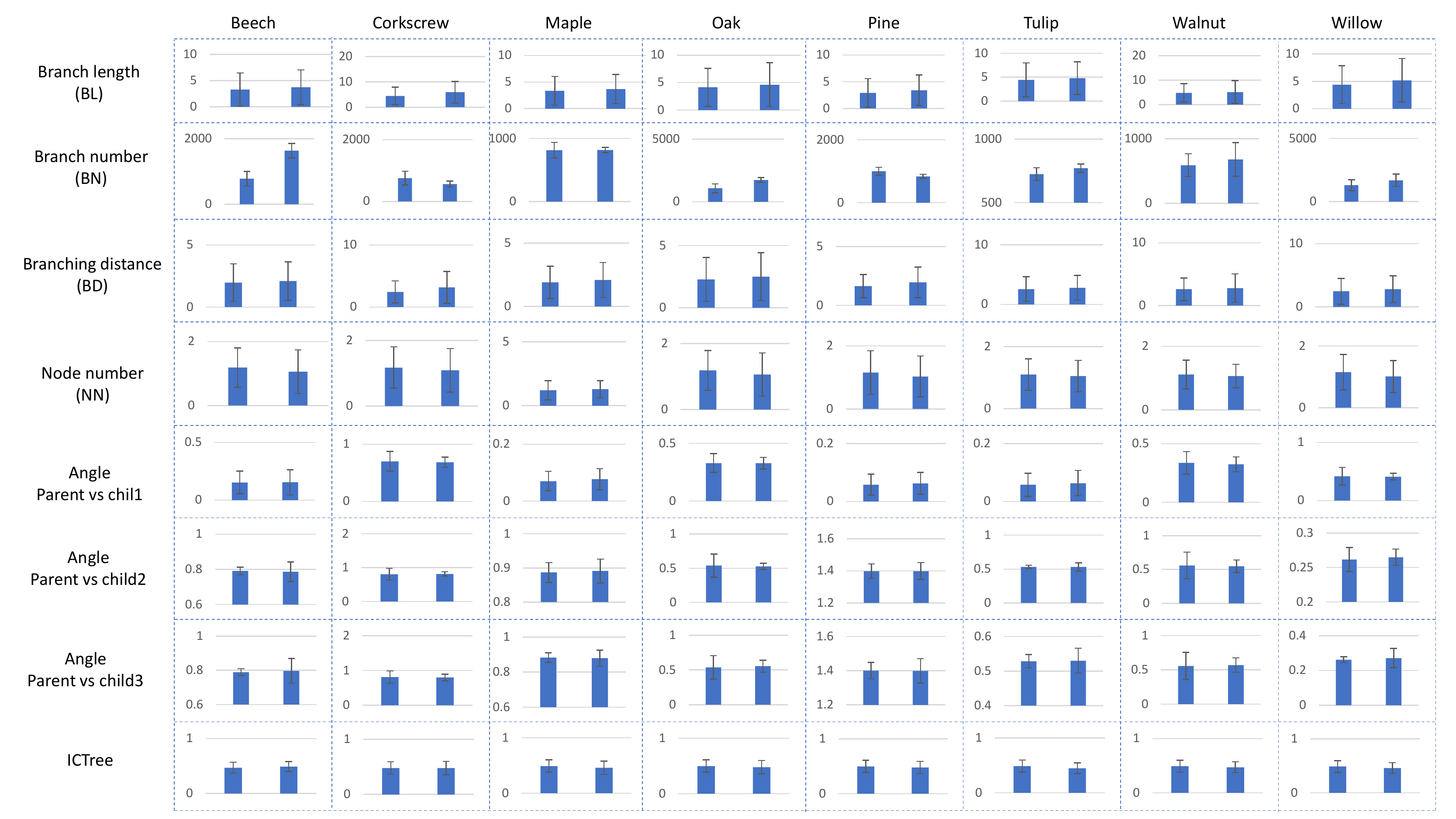}
\vspace{-3mm}
\caption{Geometric and perceptual comparison of the ground truth (GT) (left bar) and the trees generated by \name~(DT) (the right bar).}
\vspace{-3mm}
\label{fig:nodeFreq}
\end{figure*}

\paragraph{Geometry:} We validate the geometric structure of the trees generated by our \name~approach by comparing their geometric properties to the ground truth, and the results are shown in Fig.~\ref{fig:nodeFreq} and Tab.~\ref{tab:parameter_values} (Appx.). The overall branch length varies by 11\% for all trees, the number of generated branches by 18\%, branching distance by 12\%, the number of generated nodes by 11\%, and the angles (in order from the trunk) by 6\%, 4\%, and 6\%. The variations are minor, and it is essential to note that there is a great difference in visual importance for the presented features, e.g., the angle of the branches coming from the trunk has a strong effect on the overall tree shape, the branching angle of the small branchlets is not so important. This observation is further supported by the perceptual metrics.

\paragraph{Perceptual Metrics:} While the geometric comparison shows that our trees have geometric traits comparable to the ground truth, we also want to answer whether the generated trees are perceived as realistic by humans. While a standard and tedious approach to this validation is to run a user study, we validate the generated trees by recently introduced deep neural perceptual metrics ICTree~\cite{Polasek2021ToG} that estimate the visual realism of the 3D tree models. ICTree has been trained by a response of 4,000 human subjects, and it provides a value between zero to one that corresponds to the perceived visual realism of the input tree. The authors introduced view-independent ICTreeF and image-based ICTreeI metrics. We use the ICTreeF that uses geometric tree properties. 

Tab.~\ref{tab:perceptual} shows the average and the standard deviation of the perceived realism of the ground truth and the generated trees. The overall perceived realism metrics of all generated trees is $0.47\pm{0.11}$ that is 4.3\% different from the ground truth of $0.49\pm{0.11}$.
\begin{table}[t]
\caption{ICTree perceptual validation of the ground truth and generated trees (mean and std for the entire dataset).}
\vspace{-3mm}
    \centering
    \scalebox{0.85}{
    \begin{tabular}{c l l}
    \toprule
    Species & Ground Truth & \name\\
    \midrule
    Beech & [0.47, 0.10] & [0.49, 0.09]\\
    Corkscrew & [0.48, 0.11] & [0.48, 0.12]\\
    Maple & [0.49, 0.11] & [0.46, 0.12]\\
    Oak & [0.50, 0.11] & [0.48, 0.12]\\
    Pine & [0.50, 0.11] & [0.48, 0.11]\\
    Tulip & [0.49, 0.11] & [0.45, 0.10]\\
    Walnut & [0.49, 0.11] & [0.47, 0.10]\\
    Willow & [0.49, 0.11] & [0.46, 0.10]\\
    \midrule
    \textbf{All trees} & \textbf{[0.49, 0.11]} & \textbf{[0.47, 0.11]}\\
    \bottomrule
    \end{tabular}
    }
    \label{tab:perceptual}
\vspace{5mm}
\caption{Runtime performance comparison of \chh{procedurally generated trees~(P) obtained with the method of~\cite{Stava2014}} and \name~generated trees (DT). The runtime is reported in seconds and only includes the time required to generate the branch graph.}
\vspace{-3mm}
    \centering
    \scalebox{0.76}{
    \begin{tabular}{l c c c c }
    \toprule
    Species & Method & Runtime & \# Nodes & \# Branches\\
    \midrule
    \multirow{2}{*}{Beech}      & P   & 23 & 3376 & 777\\
                                & DT  & 2.7 & 5672 & 1634\\
    \midrule
    \multirow{2}{*}{Corkscrew}  & P   & 26 & 2872 & 751\\
                                & DT  & 2.1 & 3041 & 566\\                          
    \midrule
    \multirow{2}{*}{Maple}      & P   & 25 & 2928 & 861\\
                                & DT  & 1.9 & 2951 & 817\\
    \midrule
    \multirow{2}{*}{Oak}        & P   & 27 & 4261 & 1084\\
                                & DT  & 2.1 & 6238 & 1738\\                                
    \midrule
    \multirow{2}{*}{Pine}       & P   & 26 & 3506 & 1000\\
                                & DT  & 1.9 & 3168 & 839\\                                
    \midrule
    \multirow{2}{*}{Tulip}      & P   & 26 & 3050 & 726\\
                                & DT  & 1.8 & 3360 & 773\\         
    \midrule
    \multirow{2}{*}{Walnut}      & P   & 27 & 5638 & 589\\
                                 & DT  & 2.0 & 4975 & 677\\                                         
    \midrule
    \multirow{2}{*}{Willow}      & P   & 30 & 4796 & 1304\\
                                 & DT  & 2.0 & 4975 & 1691\\                                          
    \bottomrule
    \end{tabular}
    }
    \label{tab:runtime_performance}
    \vspace{-4mm}
\end{table}

\paragraph{Runtime Performance:} Tab.~\ref{tab:runtime_performance} shows the comparison of the runtime performance of procedurally generated tree models and \name. The reported numbers represent the average measurements of each method for ten models for each species. \ch{we report the measured time that the two algorithms take to generate the branching structure of a tree model.} As shown, our method outperforms the procedural algorithm for tree models of the same species and similar complexity. While the procedural model is implemented with a recursive algorithm to construct the branch graph, our method queries the neural network pipeline of classification and a cascade of regression networks to generate the branch graph iteratively. 

\chh{\paragraph{Network Comparison} To validate our cascaded neural network architecture (Fig.~\ref{fig:network_architecuture}c), we compare it with a simpler architecture that jointly trains the classification and regression networks. The architecture for this experiment is the same as shown in Fig.\ref{fig:network_architecuture}a, except that we add two additional heads on top of the joined embedding to generate thickness and length as well as the quaternion. We trained both architectures on all species with the same setup for loss functions and data processing. With the cascaded network architecture, we are able to obtain a classification result of 83 - 93\%, an MSE of 1.2e-4 - 3e-4, and a cosine similarity of 0.04 - 0.20, whereas, for the joint model, we obtain a classification result of 89\% - 90\%, an MSE of 0.28 - 0.51, and a cosine similarity of 1.03 - 0.51. While the classification head is trained with similar accuracy, the regression results are inferior. The jointly trained network immediately predicted erroneous branching that caused the termination of the tree generation. 
}

%% file: src/06-conclusion.tex

\section{Discussion and Limitations}
Our focus was on exploring the capabilities of neural networks to predict branching structures by only training them on local branching patterns. Training a neural network this way learns a situated latent space - a representation that can encode the necessary topological and geometric information to mimic branch growth. As we have shown, a trained situated latent space can serve as a powerful representation capable of generating a wide range of complex branching patterns and encoding the environmental response of trees to obstacles in their environment. 

Using situated latent spaces, our approach is orthogonal to approaches that aim to learn graph structures end-to-end. In contrast to these approaches, our method is driven by the idea of only encoding local distributions, e.g., local variations of branching patterns. This is inspired by Nature's ability to encode structural and behavioral properties as DNA, where the DNA serves as a blueprint for generating complex structures. \name~is a step in this direction that intriguingly shows that locally learned representations can provide an interesting modeling alternative.  

Our work is also similar to graph neural networks (GNNs) in that we also focus on predicting the topological and geometric properties of graphs. Research toward GNNs has recently gained a lot of momentum, and many approaches aim to predict graphs and their properties in different ways~\cite{ZHOU202057,9046288,XU202169,Shao:2021:GraphLearning}. GNNs leverage the connectivity of vertices to their neighbors in a graph by simultaneously predicting vertex properties and their connectivity information. Most of these methods focus on making predictions for local properties in graphs to solve classification tasks. Our work aims to simultaneously estimate the topology and the geometric properties of graph nodes. Regressing multiple geometric attributes remains a challenging problem that we addressed by training a cascade of individually trained regression networks.

Our method is currently limited in three ways: first, we rely on using data generated by a procedural model. We decided to showcase our method on procedurally generated tree models as this is the only way to generate a large dataset to train our neural network pipeline successfully. To the best of our knowledge, large collections of reconstructed models from scans or images do not exist. However, as long as the tree models are generated from branching graphs, which is also common practice for reconstruction algorithms and manual artistic modeling, our method can still work with the same degree of sophistication. \ch{Moreover, \name\ is a data-driven method, and the training data determines the set of generated trees. While we can generate variations via environmental changes, modifying the parameters of the intrinsic model is not possible.} The second limitation is that our method is currently unable to model tree-tree or branch-branch interactions. Procedural models can distinguish between the intrinsic plant environment (the plant itself) and the extrinsic (obstacles). However, as we rely on a coarse voxel space to encode the environment, we cannot solve intricate tree-tree or branch-branch collisions during tree generation. Finally, we observed that our method has difficulties correctly predicting the termination of branches for some node configurations, which generates artificial-looking branching structures (Fig.~\ref{fig:failture_cases}). While this leads to unrealistic branching structures for some tree models, we observed that this is highly correlated with the size of the dataset and the training duration. 
\begin{figure*}[hbt]
\centering
\includegraphics[width=0.99\linewidth]{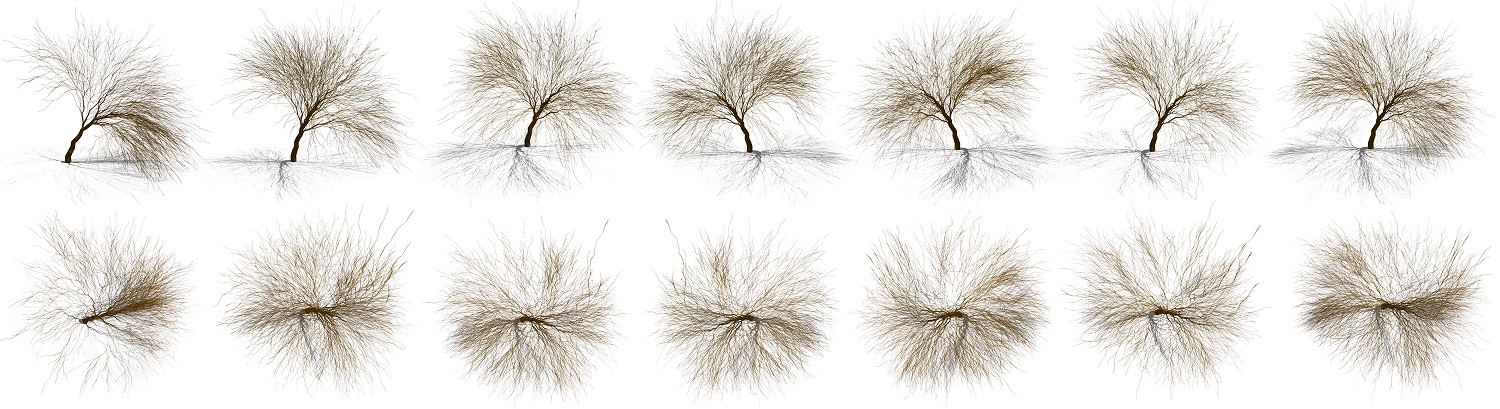}
\vspace{-5mm}
\caption{\chh{Random sampling: a set of tree models of the same species generated by our neural network pipeline shown from the front (top row) and top (bottom).}}
\vspace{-3mm}
\label{fig:stochasticity}
\end{figure*}

\section{Conclusions and Future Work}
We have advanced tree modeling in computer graphics by introducing \name, a deep-learning-based method for automatic and adaptive tree form generation. \name~is inspired by Nature's ability to encode the development of structural and behavioral features as DNA and decides the shape generation on a local level based on the signature of a single node and the environment. Instead of manually defining parameters and rules -- as is commonly done in procedural modeling -- our method learns branching patterns locally as they can be observed for a single node in a branch graph. We have shown that a situated latent space that is evaluated for a single node in a tree graph, somewhere in the growth space of a tree model, can generate complex tree models with similar topological and geometric features as contemporary procedural and developmental modeling techniques. 

We have shown that our method can generate a diverse set of branching structures from different species. Adding additional information to each node's signature (e.g., a species identifier) shows that our method can even be trained in a generalized manner, i.e., we can jointly train our networks on multiple species and successfully generate tree models. Furthermore, by encoding the environment, we have shown that our situated latent space can even support mimicking the environmental response of tree models when they grow next to walls or other trees. 

A novel way of encoding tree form opens multiple avenues for future work. First, it would be interesting to explore how a situated latent space can be used to support the authoring of branching structures. In our current framework, we only automatically use our network pipeline. However, we could also control our network pipeline by user-defined sketches or adapt it to dynamically changing scenes. Second, our current architecture does not avoid inter-branch collisions. To this end, it would be interesting to explore additional global or semi-global encodings for the generated branches when querying the networks for producing new nodes. For example, encoding the neighboring nodes of a node could provide an additional feature for producing branching structures that do not collide. It also seems promising to use our proposed algorithm of iteratively querying a neural network pipeline to generate other objects, e.g., road networks or buildings.

%% file: src/99-appendix.tex
\appendix
\newpage
\section{Appendix}

\begin{table}[h]
\vspace{-2mm}
\caption{Table of parameters for the procedural model.}
\vspace{-4mm}
\begin{center}
\scalebox{0.65}{
\begin{tabular}{ p{0.9cm}  p{3.5cm}  p{6.5cm}}
\hline
\textbf{Param} & \textbf{Name} & \textbf{Description} \\ \hline
$G_{NLB}$ & Number of lateral buds & The number of lateral buds created per internode during growth. \\ 
$G_{AAM}$ & Apical angle Mean & The mean of the angle between the direction of parent shoot and the direction of apical bud. \\ 
$G_{AAV}$ & Apical angle variance & The variance of the angle between the direction of parent shoot and the direction of apical bud. \\ 
$G_{BAM}$ & Branching angle mean & The mean of angle between the direction of parent shoot and the direction of lateral bud. \\ 
$G_{BAV}$ & Branching angle variance & The variance of angle between the direction of parent shoot and the direction of lateral bud. \\ 
$G_{RAM}$ & Roll angle mean & The mean of orientation angle between two lateral buds created with the same internode. \\
$G_{RAV}$ & Roll angle variance & The variance of orientation angle between two lateral buds created with the same internode. \\ 
$F_{AKP}$ & Apical bud kill probability & The probability that a given apical bud will die during a growth cycle. \\ 
$F_{LKP}$ & Lateral bud kill probability & The probability that a given lateral bud will die during a growth cycle. \\ 
$F_{ALF}$ & Apical bud lighting factor & The influence of the lighting condition on the growth probility of a apical bud. \\ 
$F_{LLF}$ & Lateral bud lighting factor & The influence of the lighting condition on the growth probility of a lateral bud. \\ 
$F_{ADB}$ & Apical dominance base & The base level of auxin produced to inhibit parent shoots from growing. \\ 
$F_{ADF}$ & Apical dominance distance factor & The reduction of auxin due to the transimission along parent shoots. \\ 
$F_{AAF}$ & Apical dominance age factor & The reduction of auxin due to increasing age of the tree. \\
$F_{GR}$ & Growth rate & The expected number of internodes generated along the branch during a growth cycle. \\
$F_{ILB}$ & Internode length base & The base distance between two adjacent internodes on the same shoot. \\ 
$F_{ACB}$ & Apical control base & The impact of the branch level on the growth rate. \\
$E_{PHO}$ & Phototropism & The impact of the average growth direction of incoming light. \\ 
$E_{GGB}$ & Gravitropism base & The impact of the average growth direction of the gravity. \\ 
$E_{PPF}$ & Pruning factor & The impact of the amount of incoming light on the shedding of branches. \\
$E_{LPF}$ & Low branch pruning factor & The height below which all lateral branches are pruned. \\ 
$E_{GBS}$ & Gravity bending strength & The impact of gravity on branch structural bending. \\
$E_{GAF}$ & Gravity bending angle factor & The relation of gravity bending related to the thickness of the branch. \\
t & Desired age & The expected age of the tree. \\ 
\hline
\end{tabular}
}
\end{center}
\label{tab:parameters}
 \end{table}

\begin{table}[h]
\centering
\caption{Parameter values for each species used in our framework.}
\vspace{-4mm}
\begin{center}
\scalebox{0.65}{
\begin{tabular}{lrrrrrrrr}
\hline
\textbf{Params.} & \textbf{Beech} & \textbf{Corkscrew} & \textbf{Maple} & \textbf{Oak} & \textbf{Pine} & \textbf{Tulip} & \textbf{Walnut} & \textbf{Willow}\\ \hline
$G_{NLB}$ & 2       & 2     & 2     & 2     & 2     & 2     & 2     & 2     \\ 
$G_{AAM}$ & 0       & 45    & 0     & 20    & 0     & 0     & 20    & 25    \\ 
$G_{AAV}$ & 2       & 3     & 0     & 2     & 0     & 1     & 2     & 3     \\ 
$G_{BAM}$ & 45      & 45    & 50    & 30    & 80    & 30    & 30    & 15    \\ 
$G_{BAV}$ & 1       & 2     & 1     & 3     & 1     & 1     & 3     & 1     \\ 
$G_{RAM}$ & 120     & 120   & 110   & 130   & 120   & 125   & 120   & 120   \\ 
$G_{RAV}$ & 3       & 3     & 2     & 3     & 3     & 3     & 1     & 2     \\ 
$F_{AKP}$ & 0.0     & 0.0   & 0.0   & 0.01  & 0.0   & 0.05  & 0.1   & 0.0   \\ 
$F_{LKP}$ & 0.01    & 0.03  & 0.0   & 0.03  & 0.03  & 0.21  & 0.15  & 0.12  \\ 
$F_{ALF}$ & 0.1     & 0.23  & 0.12  & 0.15  & 0.05  & 0.09  & 0.04  & 0.3   \\ 
$F_{LLF}$ & 0.54    & 1.13  & 0.94  & 0.2   & 0.2   & 1.0   & 0.1   & 0     \\ 
$F_{ADB}$ & 0.2     & 0.12  & 0.2   & 0.12  & 0.5   & 0.5   & 0.12  & 0.11  \\ 
$F_{ADF}$ & 1.0     & 1.0   & 1.0   & 0.98  & 1.0   & 1.0   & 0.1   & 1.0   \\ 
$F_{AAF}$ & 0.5     & 0.3   & 0.5   & 0.3   & 0.3   & 0.3   & 0.3   & 0.3   \\ 
$F_{GR}$ & 1.5      & 1.45  & 1.45  & 1.35  & 1.2   & 1.1   & 1.75  & 1.12  \\ 
$F_{ILB}$ & 1.0     & 1.0   & 1.0   & 1.0   & 1.0   & 1.0   & 1.0   & 1.0   \\ 
$F_{ACB}$ & 1.0     & 1.0   & 1.5   & 1.0   & 1.6   & 2.0   & 6.2   & 1.0   \\ 
$E_{PHO}$ & 0.01    & 0.1   & 0.0   & 0.05  & 0.0   & 0.1   & 0.05  & 0.1   \\ 
$E_{GGB}$ & -0.25   & -0.21 & -0.1  & -0.1  & -0.09 & -0.15 & -0.1  & 0.07  \\ 
$E_{PPF}$ & 0.1     & 0.2   & 0.05  & 0.01  & 0.01  & 0.7   & 0.12  & 0     \\ 
$E_{LPF}$ & 0.1     & 0.15  & 0.1   & 0.15  & 0.15  & 0.15  & 0.15  & 0.2   \\ 
$E_{GBS}$ & 1.0     & 6.0   & 2.0   & 6.0   & 4.0   & 2.0   & 6.0   & 0.0   \\ 
$E_{GAF}$ & 2.0     & 3.0   & 2.0   & 3.0   & 2.0   & 3.0   & 3.0   & 3.0   \\ 
t         & 35      & 35    & 40    & 35    & 45    & 60    & 40    & 40    \\ 
\hline
\end{tabular}
}
\end{center}
\label{tab:parameter_values}
\end{table}

\begin{table*}[h!]

\centering
\caption{Evaluation metrics values for each species. P indicates procedurally generated models and DT~\name~generated ones. The top row in each pair of rows shows the mean value and the second row the standard deviation.}
\label{tab:parameter_values}
\vspace{-2mm}

\scalebox{0.85}{
\begin{tabular}{l|cccccccccccccccc|cc|c}

\hline
\textbf{Params.} & \multicolumn{2}{c}{\textbf{Beech}} & \multicolumn{2}{c}{\textbf{Corkscrew}} & \multicolumn{2}{c}{\textbf{Maple}} & \multicolumn{2}{c}{\textbf{Oak}} & \multicolumn{2}{c}{\textbf{Pine}} & \multicolumn{2}{c}{\textbf{Tulip}} & \multicolumn{2}{c}{\textbf{Walnut}} & \multicolumn{2}{c}{\textbf{Willow}} &\multicolumn{2}{c}{\textbf{Total}} &$\Delta$\\ 

 &P &DT&P &DT&P &DT&P &DT&P &DT&P &DT&P &DT&P &DT&P &DT\\ 
\hline
Branch length &3.30 &3.75 &4.47 &5.92 &3.34 &3.69 &4.16 &4.61 &2.92 &3.41 &4.44 &4.80 &4.84 &5.13 &4.43 &5.19 &4.50 &4.04 &11\%\\
(BL)          &3.16 &3.31 &3.48 &4.29 &2.70 &2.75 &3.44 &3.97 &2.67 &2.88 &3.56 &3.40 &3.81 &4.67 &3.46 &3.97 &3.77 &3.36\\ \midrule

Branch number &777 &1634 &751 & 566 &816 &817 &1084 &1738 &1000 &839 &726 &773 &589 &677 &1304 &1691 &894 &1092 &18\%\\
(BN)             &222 &308 &224 &95 &122 &42 &354 &194 &129 &65 &49 &33 &175 &259 &440 &495 &337 &527\\\midrule

Branching distance &1.98 &2.10 &2.43 &3.16 &1.91 &2.11 &2.28 &2.50 &1.65 &1.96 &2.56 &2.79 &2.60 &2.79 &2.47 &2.80 &2.47 &2.80 &12\%\\
(BD)    &1.51 &1.55 &1.79 &2.57 &1.27 &1.38 &1.76 &1.92 &0.98 &1.30 &2.06 &2.08 &1.82 &2.26 &2.03 &2.14 &2.03 &2.14\\ \midrule

Node number &1.19 &1.06 &1.18 &1.09 &1.20 &1.29 &1.18 &1.06 &1.16 &1.03 &1.10 &1.05 &1.11 &1.07 &1.15 &1.01 &1.15 &1.08 &11\%\\
(NN)        &0.62 &0.68 &0.63 &0.67 &0.75 &0.68 &0.60 &0.66 &0.69 &0.65 &0.51 &0.51 &0.45 &0.37 &0.58 &0.52 &0.59 &0.59\\ \midrule

Angle: parent vs child1 &0.15 &0.15 &0.70 &0.69 &0.07 &0.08 &0.33 &0.33 &0.06 &0.06 &0.06 &0.06 &0.34 &0.32 &0.42 &0.41 &0.31 &0.29 &6\%\\
  &0.10 &0.11 &0.17 &0.09 &0.03 &0.04 &0.08 &0.05 &0.04 &0.04 &0.04 &0.04 &0.10 &0.06 &0.15 &0.06 &0.24 &0.19\\ \midrule

Angle: parent vs child2 &0.79 &0.79 &0.81 &0.81 &0.89 &0.89 &0.54 &0.53 &1.40 &1.40 &0.53 &0.53 &0.56 &0.55 &0.26 &0.27 &0.70 &0.67 &4\%\\
    &0.02 &0.06 &0.18 &0.07 &0.03 &0.04 &0.17 &0.04 &0.05 &0.05 &0.02 &0.06 &0.20 &0.10 &0.02 &0.01 &0.38 &0.31\\ \midrule

Angle: parent vs child3 &0.79 &0.80 &0.81 &0.80 &0.88 &0.88 &0.54 &0.56 &1.40 &1.40 &0.53 &0.53 &0.56 &0.57 &0.26 &0.27 &0.70 &0.66 &6\%\\
    &0.02 &0.07 &0.18 &0.10 &0.03 &0.05 &0.17 &0.09 &0.05 &0.07 &0.02 &0.04 &0.20 &0.18 &0.02 &0.06 &0.38 &0.32\\ \bottomrule
\end{tabular}
}
\end{table*}